\documentclass{article}
\usepackage{fullpage}

\usepackage{blindtext}
 
\usepackage{xcolor}
\usepackage[utf8]{inputenc} % allow utf-8 input
\usepackage{mysymbol}

\usepackage[T1]{fontenc}    % use 8-bit T1 fonts
\usepackage{hyperref}       % hyperlinks
\usepackage{url}            % simple URL typesetting
\usepackage{booktabs}   
\usepackage{color}
\usepackage{amsfonts}       % blackboard math symbols
\usepackage{nicefrac}       % compact symbols for 1/2, etc.
\usepackage{microtype}      % microtypography
\usepackage{lipsum}
\usepackage{fancyhdr}       % header
\usepackage{graphicx}       % graphics
\graphicspath{{media/}}     % organize your images and other figures under media/ folder
\usepackage{enumitem}
\usepackage{amsthm}

\usepackage[utf8]{inputenc} % allow utf-8 input
\usepackage[T1]{fontenc}    % use 8-bit T1 fonts
\usepackage{url}            % simple URL typesetting
\usepackage{booktabs}       % professional-quality tables
\usepackage{amsfonts}       % blackboard math symbols
\usepackage{nicefrac}       % compact symbols for 1/2, etc.
\usepackage{microtype}      % microtypography

\usepackage{algorithm}
\usepackage{longtable}

\newtheorem{mydef}{Definition}
\newtheorem{theorem}{Theorem}

\newtheorem{lemma}{Lemma}

\newtheorem{prop}{Proposition}
\newtheorem{assumption}{Assumption}

\usepackage[utf8]{inputenc}
\usepackage{pgfplots}
\usepackage{graphicx}
\usepackage{subcaption}
\usepackage{pstricks}
\usepackage{color}
\usepackage{amsmath}

\allowdisplaybreaks

\usepackage{bbm}
\usepackage{tcolorbox}
\usepackage{tikz}
\usepackage{pgfplots}
\usepackage{wrapfig}
\usepackage{lipsum}  
\usetikzlibrary{positioning}
\usepackage{tcolorbox}
\usepackage{amsfonts,amssymb}
\usepackage{mathtools}
\usepackage{commath}
\usepackage{relsize}
\usepackage{bbm}
\usepackage{bm}
\usepackage[font={small}]{caption}
\usepackage{comment,color,soul}
\usetikzlibrary{arrows,automata}
\usepackage{amsfonts}
\usepackage{url}
\usepackage{lipsum}
\usepackage[thinlines]{easytable}
\usepackage{nicefrac}
\usepackage{algorithm}
\usepackage{algpseudocode}
\usepackage{float}
\usepackage{multirow}
\usepackage{colortbl}
\usepackage[title]{appendix}
\newcommand\numberthis{\addtocounter{equation}{1}\tag{\theequation}}

\title{\textbf{On the Role of Generalization in Transferability of Adversarial Examples}}

\date{}

\author{
Yilin~Wang\thanks{Department of Computer Science and Engineering, The Chinese University of Hong Kong, 1155166083@link.cuhk.edu.hk.} , 
Farzan~Farnia\thanks{Department of Computer Science and Engineering, The Chinese University of Hong Kong, farnia@cse.cuhk.edu.hk.}
	}

\begin{document}
\maketitle

% keywords can be removed
% \keywords{First keyword \and Second keyword \and More}

\begin{abstract}
Black-box adversarial attacks designing adversarial examples for unseen neural networks (NNs) have received great attention over the past years. While several successful black-box attack schemes have been proposed in the literature, the underlying factors driving the transferability of black-box adversarial examples still lack a thorough understanding. In this paper, we aim to demonstrate the role of the generalization properties of the substitute classifier used for generating adversarial examples in the transferability of the attack scheme to unobserved NN classifiers. To do this, we apply the max-min adversarial example game framework and show the importance of the generalization properties of the substitute NN in the success of the black-box attack scheme in application to different NN classifiers. We prove theoretical generalization bounds on the difference between the attack transferability rates on training and test samples. Our bounds suggest that a substitute NN with better generalization behavior could result in more transferable adversarial examples. In addition, we show that standard operator norm-based regularization methods could improve the transferability of the designed adversarial examples. We support our theoretical results by performing several numerical experiments showing the role of the substitute network's generalization in generating transferable adversarial examples. Our empirical results indicate the power of Lipschitz regularization methods in improving the transferability of adversarial examples.        
\end{abstract}

% \vspace{-.1cm}
\section{Introduction}
Deep neural networks (DNNs) have attained impressive results in many machine learning problems from image recognition \cite{krizhevsky2009learning}, speech processing \cite{deng2013recent}, and bioinformatics \cite{alipanahi2015predicting}. The standard evaluation of a trained DNN machine is typically performed over test samples drawn from the same underlying distribution that has generated the empirical training data. The numerous successful applications of deep learning models reported in the literature  demonstrate DNNs' surprising generalization power from training samples to unseen test data. Such promising results on unobserved data despite DNNs' enormous capacity for memorizing training examples have attracted a lot of attention in the machine learning community.

While DNNs usually achieve satisfactory generalization performance, they have been frequently observed to lack robustness against minor adversarial perturbations to their input data \cite{szegedy2013intriguing,biggio2013evasion,goodfellow2014explaining}, widely known as adversarial attacks. According to these observations, an adversarial attack scheme can generate imperceptible perturbations that fools the DNN classifier to predict wrong labels with high confidence scores. Such adversarial perturbations are usually created through maximizing a target DNN's prediction loss over a small neighborhood around an input sample. While DNNs often show successful generalization behavior to test samples drawn from the underlying distribution of training data, the minor perturbations designed by adversarial attack schemes can completely undermine their prediction results.    

Specifically, adversarial examples have been commonly reported to be capable of transferring to unseen DNN classifiers \cite{tramer2017ensemble,ilyas2018black,cheng2018query,zhou2018transferable}. Based on these reports, an adversarial example designed for a specific classifier could further alter the prediction of another DNN machine with a different architecture and training set. Such observations have inspired the development of several \emph{black-box adversarial attack schemes} in which the adversarial examples are designed for a substitute classifier and then are evaluated on a different target DNN.  %However, the underlying factors contributing to the success of black-box attacks are still not completely understood. 

Several recent papers have attempted to theoretically study the transferability of black-box adversarial attacks. These works have mostly focused on the effects of non-robust features \cite{tramer2017space,ilyas2019adversarial,inkawhich2019feature} and equilibrium \cite{bose2020adversarial,meunier2021mixed} in adversarial training problems on transferable adversarial examples. The mentioned studies reveal the dependency of adversarial examples on non-robust features that can be easily perturbed through minor adversarial noise, and also how the transferability of adversarial examples depend on the equilibrium in the game between the adversary and classifier players. On the other hand, the connections between the generalization behavior of the substitute network and the transferability of the designed examples have not been explored in the literature. Therefore, it remains unclear whether a substitute DNN with better generalization performance can result in more transferable adversarial attacks.

In this work, we attempt to understand the interconnections between the generalization and attack transferability properties of DNNs in black-box adversarial attacks. We aim to show that better generalization performance not only can improve the classification accuracy on unseen data, but further could result in higher transferability rates for the designed adversarial examples. To this end, we analyze the transferability of adversarial examples through the lens of the max-min framework of \emph{adversarial example game} \cite{bose2020adversarial}. According to this approach, the adversary player searches for the most transferable attack strategy that reaches the maximum prediction error under the most robust DNN classifier. We focus on the generalization aspect of the adversarial example game, and demonstrate its importance in the transferability power of adversarial perturbations.   

Specifically, we focus on the class of norm-bounded adversarial attacks and define the generalization error of a function class's minimum risk under standard norm-bounded adversarial perturbations. Subsequently, we prove theoretical bounds on the defined generalization metric for multi-layer DNNs with spectrally-normalized weight matrices. Our result extends the operator norm-based generalization bounds \cite{bartlett2017spectrally,neyshabur2017pac,wei2019improved} in the deep learning literature to the adversarial example game, which enables us to bound the generalization error for the transferability performance of norm-bounded adversarial attack strategies. Also, the shown generalization bound suggests the application of Lipschitz regularization methods in training the substitute DNN in order to improve the transferability of designed adversarial examples.

Finally, we numerically evaluate our theoretical results on multiple standard image recognition datasets and DNN architectures. Our empirical results further support the existing connections between the generalization and transferability properties of black-box adversarial attacks. The numerical findings demonstrate that a better generalization score for the substitute DNN could significantly boost the transferability rate of designed adversarial examples. In addition, we empirically demonstrate that both explicit and implicit regularization techniques can help generating more transferable examples. We validate this result for explicit Lipschitz regularization and implicit early-stopping schemes. We can summarize the main contributions of our work as follows:
\begin{itemize}[leftmargin=7mm]
    \item Drawing connections between the generalization properties of the substitute DNN classifier and the transferability rate of designed adversarial examples
    \item Proving generalization error bounds on the difference between the transferability rates of DNN-based adversarial examples designed for training and test data
    \item Demonstrating the power of Lipschitz regularization and early stopping methods in generating more transferable adversarial examples
    \item Conducting numerical experiments on the generalization and transferability aspects of black-box adversarial attacks
\end{itemize}

\section{Related Work}
Transferability of adversarial examples has been extensively studied in the deep learning literature. The related literature includes a large body of papers \cite{ilyas2018black,cheng2018query,bhagoji2018practical,alzantot2019genattack,cheng2019improving,moon2019parsimonious,guo2019simple,mohaghegh2020advflow,wang2020amora} proposing black-box adversarial attack schemes aiming to transfer from a source DNN to an unseen target DNN classifier and several related works \cite{levine2020robustness,salman2020denoised,singla2020second,li2020blacklight} on developing robust training mechanisms against black-box adversarial attacks. Also, several game theoretic frameworks have been proposed to analyze the transferability of adversarial examples. The related works \cite{bose2020adversarial,meunier2021mixed} study the adversarial example game between the classifier and adversary players. However, these works mostly focus on the equilibrium and convergence behavior in adversarial example games and do not discuss the generalization aspect of the game. In another related work, \cite{pal2020game} studies the adversarial learning task through the lens of game theory. Unlike our work, the generalization analysis in \cite{pal2020game} focuses only on the generalization behavior of the robust classification rule and not on the generalization properties of the transferable adversary player.    

Furthermore, the generalization properties of adversarially-learned models have been the topic of several related papers.
References \cite{schmidt2018adversarially,raghunathan2019adversarial} discuss numerical and theoretical results that generalization of adversarially-trained neural nets is inferior to that of standard ERM-learned models with the same number of training data. The related work \cite{rice2020overfitting} empirically studies the overfitting phenomenon in adversarial training problems and reveals the different generalization properties of standard and adversarial training schemes. 
In another study, \cite{wu2020adversarial} shows the connection between the generalization of adversarially-learned models and the flatness of weight loss landscape. \cite{yin2019rademacher,awasthi2020adversarial} develop Rademacher-complexity-based generalization bounds for adversarially-trained models which suggest the application of norm-based regularization techniques for improving the generalization behavior of adversarial training methods. \cite{farnia2018generalizable} proves Pac-Bayes generalization bounds for adversarially-learned DNNs with bounded spectral norms for their weight matrices. Also, \cite{attias2019improved} performs VC-based generalization analysis for adversarial training schemes and derives upper-bounds on their sample complexity. However, we note that all these papers focus on the generalization of adversarially-trained models and do not study the connection between generalization and transferability of black-box attacks.  

%Finally, we note that Lipschitz regularization techniques have been applied to improve adversarial robustness and generalization of neural network models. 

\iffalse
\textcolor{red}{Some related work that might be worth discussing:
\begin{itemize}
    \item A Game Theoretic Analysis of Additive Adversarial Attacks and Defenses
    \item Adversarial Example Games
    \item Based on Max-Min Framework Transferable Adversarial Attacks
    \item Mixed Nash Equilibria in the Adversarial Examples Game
    \item Generalizable Adversarial Training via Spectral Normalization
\end{itemize}}
\fi

%\section{Preliminaries}
%\input{2-preliminaries.tex}

% \vspace{-.1cm}
\section{Preliminaries: Adversarial Attacks and Training}
In this section, we give a brief review of standard norm-bounded adversarial attack and training schemes. Consider a supervised learning problem where the learner seeks a prediction rule $f$ from  function space $\mathcal{F}$ to predict a label variable $Y\in\mathcal{Y}$ from the observation of a $d$-dimensional feature vector $\mathbf{X}\in\mathcal{X}$. In this work, we focus on the following set of $L$-layer neural network functions with activation function $\phi$:
\begin{equation}
    \mathcal{F_W}\, = \,\left\{\, f_\mathbf{w}:\; f_\mathbf{w}(\mathbf{x})=W_L \phi\bigl(W_{L-1}\phi(\cdots W_1\phi(W_{0}\mathbf{x}) \cdot\bigr),\, \mathbf{w}\in\mathcal{W}  \right\}.
    \label{neural_network_functions}
\end{equation}
In the above, we use vector $\mathbf{w}$ belonging to  feasible set $\mathcal{W}$ to parameterize the $L$-layer neural net $f_{\mathbf{w}}$. According to this notation,  $\mathbf{w}$ concatenates all the entries of the neural net's weight matrices $W_0,\ldots,W_L$.

Given a loss function $\ell$ and $n$ training samples in dataset $S=\{(\mathbf{x}_i,y_i)_{i=1}^n\}$, the standard risk minimization approach aims to find the prediction rule $f^*\in \mathcal{F_W}$ minimizing the expected loss (risk) $\mathbb{E}[\ell(f(\mathbf{X}),Y)]$ where the expectation is taken according to the underlying distribution of data $P_{\mathbf{X},Y}$. Since the supervised learner only observes the training samples and lacks any further knowledge of the underlying $P_{\mathbf{X},Y}$, the empirical risk minimization (ERM) framework sets out to minimize the empirical risk function estimated using the training examples:
\begin{equation}\label{Eq: ERM Problem}
    \min_{\mathbf{w}\in\mathcal{W}}\: \frac{1}{n}\sum_{i=1}^n \ell\bigl(f_\mathbf{w}(\mathbf{x}_i),y_i\bigr).
\end{equation}
However, the ERM learner typically lacks robustness to norm-bounded adversarial perturbations. A standard approach to generate a norm-bounded adversarial perturbation is through maximizing the loss function over a norm ball around a given data point $(\mathbf{x},y)$:
\begin{equation}\label{Eq: Standard Perturbation}
    \max_{\boldsymbol{\delta}:\: \Vert \boldsymbol{\delta}\Vert\le \epsilon}\; \ell\bigl(f(\mathbf{x}+\boldsymbol{\delta}),y\bigr).
\end{equation}
Here $\boldsymbol{\delta}\in\mathbb{R}^d$ represents the $d$-dimensional perturbation vector added to the feature vector $\mathbf{x}$, and $\Vert\cdot \Vert$ denotes a norm function used to measure the attack power that is bounded by parameter $\epsilon\ge 0$. 

In order to gain robustness against norm-bounded perturbations, the adversarial training (AT) scheme \cite{madry2017towards} alters the ERM objective function to the expected worst-case loss function over norm-bounded adversarial perturbations and solves the following min-max optimization problem:
\begin{equation}\label{Eq: AT Empirical distribution}
     \min_{\mathbf{w}\in\mathcal{W}}\: \frac{1}{n}\sum_{i=1}^n\biggl[ \max_{\boldsymbol{\delta}_i:\: \Vert \boldsymbol{\delta}_i\Vert\le \epsilon}\: \ell\bigl(f_\mathbf{w}(\mathbf{x}_i+\boldsymbol{\delta}_i),y_i\bigr)\biggr] \: \equiv \: \min_{\mathbf{w}\in\mathcal{W}}\: \max_{\substack{\boldsymbol{\delta}_1,\ldots,\boldsymbol{\delta}_n: \\
   \forall i, \: \Vert \boldsymbol{\delta}_i \Vert\le \epsilon}
   }\: \frac{1}{n}\sum_{i=1}^n\bigl[ \ell\bigl(f_\mathbf{w}(\mathbf{x}_i+\boldsymbol{\delta}_i),y_i\bigr)\bigr]
\end{equation}
Note that the above minimax problem indeed estimates the solution to the following learning problem formulated over the true distribution of data $P_{\mathbf{X},Y}$:
\begin{equation}\label{Eq: AT True distribution}
     \min_{\mathbf{w}\in\mathcal{W}}\; \mathbb{E}_{(\mathbf{X},Y)\sim P}\biggl[\, \max_{\boldsymbol{\delta}:\: \Vert \boldsymbol{\delta}\Vert\le \epsilon}\: \ell\bigl(f_\mathbf{w}(\mathbf{X}+\boldsymbol{\delta}),Y\bigr)\biggl].
\end{equation}
It can be seen that the above optimization problem is indeed equivalent to the following min-max problem where the maximization is performed over $\Delta_\epsilon $ containing all mappings $\delta:\mathcal{X}\times \mathcal{Y} \rightarrow \mathbb{R}^d$ whose output is $\epsilon$-norm-bounded, i.e. $\forall \mathbf{x},y: \: \Vert \delta(\mathbf{x},y)\Vert\le \epsilon$:
\begin{equation}\label{Eq: AT True distribution_MinMax}
     \min_{\mathbf{w}\in\mathcal{W}}\; \max_{\delta \in \Delta_\epsilon}\; \mathbb{E}_{\mathbf{X},Y\sim P}\bigl[\, \ell\bigl(f_\mathbf{w}(\mathbf{X}+\delta(\mathbf{X},Y)),Y\bigr)\bigl].
\end{equation}
In next sections, we will discuss the association between the above min-max problem and the adversarial example game for generating transferable adversarial examples.

% \vspace{-.1cm}
\section{A Max-Min Approach to Transferable Adversarial Examples}
Transferability of adversarial examples has been extensively studied in the literature. A useful framework to theoretically study transferable examples is the max-min framework of \emph{adversarial example game (AEG)} \cite{bose2020adversarial}. According to this approach, the adversary searches for the most transferable attack scheme $\delta\in\Delta$ from a set of attack strategies $\Delta$ that achieves the maximum expected loss under the most robust classifier $f_{\mathbf{w}}\in\mathcal{F_W}$ from DNN function space $\mathcal{F_W}$. Therefore, the AEG approach reduces the transferable adversary's task to solving the following max-min optimization problem:
\begin{equation}\label{Eq: AEG empirical distribution}
     \max_{\delta \in \Delta}\; \min_{\mathbf{w}\in\mathcal{W}}\; \frac{1}{n}\sum_{i=1}^n\biggl[ \ell\bigl(f_\mathbf{w}(\mathbf{x}_i+\delta(\mathbf{x}_i,y_i)),y_i\bigr) \biggr] %\; \approx \; \max_{\delta \in \Delta_\epsilon}\; \min_{\mathbf{w}\in\mathcal{W}}\;  \mathbb{E}_{\mathbf{X},Y\sim P}\bigl[\, \ell\bigl(f_\mathbf{w}(\mathbf{X}+\delta(\mathbf{X},Y)),Y\bigr)\bigl].
\end{equation}
The above bi-level optimization problem indeed swaps the maximization and minimization order of the AT optimization problem, and focuses on the max-min version of the min-max AT optimization task. Note that as shown in \cite{meunier2021mixed}, the adversarial example game is in general not guaranteed to possess a pure Nash equilibrium where each player's deterministic strategy is optimal when fixing the other player's strategy. As a result of the lack of pure Nash equilibria, the AEG max-min and AT min-max optimization problems may not share any common solutions. 

Note that the AEG framework introduces the following metric for evaluating the transferability of an attack scheme $\delta: \mathcal{X}\times\mathcal{Y}\rightarrow \mathbb{R}^d$:
 \begin{equation}\label{Eq: Transferability def empirical distribution}
    \widehat{\mathcal{L}}_{\text{\rm transfer}}(\delta) \, :=\, \min_{\mathbf{w}\in\mathcal{W}}\; \frac{1}{n}\sum_{i=1}^n\bigl[ \ell\bigl(f_\mathbf{w}(\mathbf{x}_i+\delta(\mathbf{x}_i,y_i)),y_i\bigr) \bigr] %\; \approx \; \max_{\delta \in \Delta_\epsilon}\; \min_{\mathbf{w}\in\mathcal{W}}\;  \mathbb{E}_{\mathbf{X},Y\sim P}\bigl[\, \ell\bigl(f_\mathbf{w}(\mathbf{X}+\delta(\mathbf{X},Y)),Y\bigr)\bigl].
\end{equation}
The above transferability score indeed estimates the following score measuring transferability under the underlying distribution $P_{\mathbf{X},Y}$:  
\begin{equation}\label{Eq: Transferability def true distribution}
    \mathcal{L}_{\text{\rm transfer}}(\delta)\, := \, \min_{\mathbf{w}\in\mathcal{W}}\; \mathbb{E}_{(\mathbf{X},Y)\sim P}\biggl[ \ell\left(f_\mathbf{w}(\mathbf{X}+\delta(\mathbf{X},Y)),Y\right)\biggr].
\end{equation}
Based on this discussion, the AEG optimization problem in \eqref{Eq: AEG empirical distribution} similarly estimates the solution to the following max-min AEG problem formed around the underlying distribution $P_{\mathbf{X},Y}$:
\begin{equation}\label{Eq: AEG true distribution}
      \max_{\delta \in \Delta}\:\mathcal{L}_{\text{\rm transfer}}(\delta) \; \equiv \; \max_{\delta \in \Delta}\; \min_{\mathbf{w}\in\mathcal{W}}\;  \mathbb{E}_{(\mathbf{X},Y)\sim P}\biggl[ \ell\left(f_\mathbf{w}(\mathbf{X}+\delta(\mathbf{X},Y)),Y\right)\biggr].
\end{equation}
Therefore, the primary goal of the transferable adversary is to solve the above problem targeting the distribution of test data instead of training examples. However, since the true distribution is unknown to the adversary, the AEG framework switches to the empirical max-min problem \eqref{Eq: AEG empirical distribution}. This discussion motivates the following definition of the generalization error for adversarial examples' transferability performance:
\begin{mydef}\label{Def :Generalization Error black-box}
We define the generalization error of an attack scheme $\delta: \mathcal{X}\times\mathcal{Y}\rightarrow \mathbb{R}^d$ over DNN classifier space $\mathcal{F_W}$ as follows:
% \begin{equation}
% \label{Eq: Gen erro black-box attack}
\begin{align}\label{Eq: Gen erro black-box attack}
    \epsilon_{\text{\rm gen}}(\delta) :=& \, \widehat{\mathcal{L}}_{\text{\rm transfer}}(\delta) -  \mathcal{L}_{\text{\rm transfer}}(\delta)\numberthis \\
    =& \,  \min_{\mathbf{w}\in\mathcal{W}}\biggl\{ \frac{1}{n}\sum_{i=1}^n\left[ \ell\bigl(f_\mathbf{w}(\mathbf{x}_i+\delta(\mathbf{x}_i,y_i)),y_i\bigr) \right]\biggr\} -  \min_{\mathbf{w}\in\mathcal{W}}\biggl\{ \mathbb{E}\left[ \ell\bigl(f_\mathbf{w}(\mathbf{X}+\delta(\mathbf{X},Y)),Y\bigr)\right] \biggr\}. \nonumber
\end{align}
% \end{equation}

\end{mydef}
In order for a black-box adversarial attack to be effective, we need the attack scheme to generalize well from training samples to test data, and based on the max-min AEG framework the generalization error is defined in the specific sense of Definition \ref{Def :Generalization Error black-box}. In next sections, we study how the choice of substitute DNN will affect the above generalization error of the resulting adversarial attack scheme. %In next section, we attempt to address this question for the class of Lipschitz DNNs.

% \vspace{-.14cm}
\section{A Generalization Error Bound for Adversarial Example Games}
In this section, we aim to analyze the generalization error of a black-box adversarial attack scheme based on the substitute classifier of a $L$-layer DNN $\mathcal{H_W}$. To characterize a one-to-one correspondence between the choice of the DNN weights and the assigned attack scheme, we consider the following definition of an optimal attack scheme for a substitute neural net $h_{\mathbf{w}}\in \mathcal{H_W}$.
\begin{mydef}
Given a classifier $h_{\mathbf{w}}$, we call the attack scheme $\delta^*_\mathbf{w}:\mathcal{X}\times \mathcal{Y}\rightarrow \mathbb{R}^d$ $\lambda$-optimal if it solves the following optimization problem:
\begin{equation*}
    \max_{\delta:\mathcal{X}\times \mathcal{Y}\rightarrow \mathbb{R}^d}\; \mathbb{E}\bigl[\ell\bigl(h_\mathbf{w}(\mathbf{X}+\delta(\mathbf{X},Y)),Y\bigr)\bigl] - \frac{\lambda}{2} \mathbb{E}\bigl[\Vert  \delta(\mathbf{X},Y) \Vert^2\bigr].
\end{equation*}
\end{mydef}
Note that the above definition of optimal attack scheme employs a regularization term to  penalize the averaged norm-squared of designed perturbations. As shown in the next proposition, this definition results in a one-to-one correspondence between $\lambda$-optimal attack schemes and $\lambda$-smooth DNN classifiers. We defer the proof of the theoretical results to the Appendix.
\begin{prop}\label{Prop: correspondence}
Consider the $L_2$-norm function $\Vert\cdot\Vert_2$ for measuring the attack power. Suppose that the composition $\ell \circ h_{\mathbf{w}}$ is a $\lambda$-smooth differentiable function of $\mathbf{x}$, i.e. for every $\mathbf{x},\mathbf{x}',y$ we have $\Vert\nabla_{\mathbf{x}}\ell(h_\mathbf{w}(\mathbf{x}),y)- \nabla_{\mathbf{x}}\ell(h_\mathbf{w}(\mathbf{x}'),y)\Vert_2\le \lambda\Vert\mathbf{x}-\mathbf{x}'\Vert_2$. Then, there exists a unique $\lambda$-optimal attack scheme $ \delta^*(\mathbf{x},y)$ for $h_{\mathbf{w}}$  given by:
\begin{equation*}
    \delta^*(\mathbf{x},y) \, =\,   \left(\text{\rm Id}_{\mathbf{x}} - \frac{1}{\lambda}\nabla_{\mathbf{x}}\ell\circ h_\mathbf{w}\right)^{-1}(\mathbf{x},y) \, - \, \mathbf{x}.
\end{equation*}
In the above equation $\text{\rm Id}_{\mathbf{x}}$ represents the identity function on feature vector $\mathbf{x}$, and $(\cdot)^{-1}$ denotes the inverse of an invertible transformation.
\end{prop}
The above proposition reveals a bijection between smooth DNN classifiers and optimal attack schemes. Therefore, in our generalization analysis, we focus on bounding the generalization error for the resulting $\lambda$-optimal attack schemes corresponding to $\lambda$-smooth DNN substitute classifiers.

In the following theorem, we show a generalization error bound for the class of $\lambda$-optimal black-box attack schemes coming from spectrally-regularized DNN functions. We note that our proof of the theorem extends the generalization analysis of spectrally-normalized neural networks in \cite{bartlett2017spectrally} to the adversarial settings which is different from the existing Rademacher complexity-based and Pac-Bayes-based generalization analyses in the related works \cite{yin2019rademacher,awasthi2020adversarial,farnia2018generalizable} on generalization of adversarial training methods. In the theorem, we use the following set of assumptions on the loss function $\ell$ and the target and substitute classes of neural networks. Also, note that $\Vert\cdot\Vert_2$ denotes the $L_2$-operator (spectral) norm in application to a matrix, i.e. the matrix's maximum singular value, and $\Vert\cdot\Vert_{2,1}$ denotes the $(2,1)$-norm of a matrix which is the summation of the $L_2$-norms of the matrix's rows. 
\begin{assumption}\label{Assumption: loss}
Loss function $\ell(y,y')$ is a $c$-bounded, $1$-Lipschitz, and $1$-smooth function of the input $y$, i.e. for every $y_1,y_2,y'\in\mathcal{Y}$ we have $\vert \ell(y_1,y')\vert\le c$, $|\ell(y_1,y')-\ell(y_2,y')| \le \Vert y_1-y_2 \Vert_2 $, and $\Vert\nabla_y\ell(y_1,y')-\nabla_y\ell(y_2,y')\Vert_2 \le \Vert y_1-y_2 \Vert_2 $.
\end{assumption}
\begin{assumption}\label{Assumption: substitute neural network}
The set of substitute DNNs in the black-box attack scheme $\mathcal{H_W}=\{h_\mathbf{w}:\, \mathbf{w}\in\mathcal{W} \}$ contains $L$-layer neural networks $h_\mathbf{w}(\mathbf{x})=W_L \phi_{L}\bigl(W_{L-1}\phi_{L-1}(\cdots W_1\phi_{1}(W_{0}\mathbf{x}) \cdot\bigr)$. We suppose that the dimensions of matrices $W_0,\ldots,W_k$ is bounded by $D$, and assume every activation $\phi_i$ satisfies $\phi_i(0)=0$ and is $\gamma_i$-Lipschitz and $\gamma_i$-smooth, i.e. $\max\{|\phi_i'(z)|,|\phi_i''(z)|\}\le \gamma_i$ holds for every $z\in\mathbb{R}$.
\end{assumption}
\begin{assumption}\label{Assumption: target neural network}
The class of target classifiers $\mathcal{F_V}=\{f_\mathbf{v}:\, \mathbf{v}\in\mathcal{V} \}$ consists of $K$-layer neural network functions $f_\mathbf{v}(\mathbf{x})=V_K \psi_{L}\bigl(V_{L-1}\psi_{L-1}(\cdots V_1\psi_{1}(V_{0}\mathbf{x}) \cdot\bigr)$ with activation function $\psi_i$'s. We suppose that the dimensions of matrices $V_0,\ldots,V_k$ is bounded by $D$. Also, we assume every $\psi_i$ satisfies $\psi_i(0)=0$ and is $\xi_i$-Lipschitz, i.e. $\max_z\: |\psi_i'(z)|\le \xi_i$. Also, we define the capacity $R_\mathcal{V}$ as
\begin{equation*}
    R_{\mathcal{V}} := \sup_{\mathbf{v}\in\mathcal{V}}\,\left\{  \bigl(\prod_{i=0}^K \xi_i \Vert V_i \Vert_2 \bigr)\biggl(\sum_{i=0}^K \frac{\Vert V^\top_i \Vert^{2/3}_{2,1}}{\Vert V_i \Vert^{2/3}_{2}} \biggr)^{3/2}\right\}.
\end{equation*}
\end{assumption}

\begin{theorem}\label{Thm: generalization bound}
Suppose that the loss function, substitute DNN, and target DNN in a black-box adversarial attack satisfy Assumptions \ref{Assumption: loss}, \ref{Assumption: substitute neural network} and \ref{Assumption: target neural network}. Assuming $\Vert \mathbf{X}\Vert_2\le B$ for the $n\times d$ data matrix $\mathbf{X}$ and $\lambda(1-\tau) \ge(\prod_{i=0}^L \gamma_i \Vert W_i \Vert_2)\sum_{i=0}^L\prod_{j=0}^L \gamma_j \Vert W_j \Vert_2 $ holds for constant $\tau > 0$ and every $\mathbf{w}\in\mathcal{W}$, then for every $\omega>0$ with probability at least $1-\omega$ the following bound will hold for every $\mathbf{w}\in\mathcal{W}$:
\begin{align}
    \epsilon_{\text{\rm gen}}(\delta^*_{\mathbf{w}})  \le \mathcal{O}\biggl(c\sqrt{\frac{\log(1/\omega)}{n}} + \frac{(B+\frac{L_{\mathbf{w}}}{\lambda}) \bigl(R_{\mathcal{V}}+\frac{1}{\tau^2}L_{\mathbf{w}}R_{\mathbf{w}}\bigr) \log(n) }{n}\log(D) \biggr) 
\end{align}
where the Lipschitz and capacity terms $L_{\mathbf{w}},\, R_{\mathbf{w}}$ are defined as:
\begin{equation}
    L_{\mathbf{w}} :=  \prod_{i=0}^L \gamma_i \Vert W_i \Vert_2 ,\quad R_{\mathbf{w}} :=  \left(\sum_{i=0}^L\prod_{j=0}^i \gamma_j \Vert W_j \Vert_2 \right)\biggl(\sum_{i=0}^L \frac{\Vert W^\top_i \Vert^{2/3}_{2,1}}{\Vert W_i \Vert^{2/3}_{2}} \biggr)^{3/2}.
\end{equation}
\end{theorem}
The above theorem bounds the generalization error of the attack scheme $\delta^*_{\mathbf{w}}$ corresponding to the substitute DNN $f_{\mathbf{w}}$ in terms of the spectral capacity of the substitute network. As a result, this bound motivates norm-based spectral regularization \cite{yoshida2017spectral,miyato2018spectral,farnia2018generalizable} for improving the generalization performance of black-box attack schemes. 

% \vspace{-.14cm}
\section{Numerical Results}
In this section, we provide the results of our numerical experiments for validating the connection between the generalization and transferability properties of black-box adversarial attacks. The numerical discussion focuses on the question of whether achieving a better generalization score for the substitute DNN can improve the success of the designed perturbations in application to a different DNN classifier. To answer this question, we tested an explicit norm-based regularization method, spectral normalization \cite{yoshida2017spectral,farnia2018generalizable}, as well as an implicit regularization technique, early stopping \cite{yao2007early,rice2020overfitting}, to evaluate the power of these regularization methods in attaining more transferable black-box attacks.

For generating norm-bounded perturbations, we used standard projected gradient descent (PGD) and fast gradient method (FGM) \cite{FGSM} to design perturbations. We implemented the PGD and FGM algorithms by projecting the perturbations according to both standard $L_2$-norm and $L_\infty$-norm, where the latter results in the widely-used fast gradient sign method (FGSM) attack scheme \cite{FGSM} in the FGM case. For simulating $L_2$-norm-bounded perturbations, we chose the maximum $L_2$-norm (attack power) as $\epsilon=\gamma\mathbb{E}_{\hat{P}}[\Vert X\Vert_2]$ with $\gamma=0.05$ unless stated otherwise. For $L_\infty$-norm-bounded attacks, we chose $\epsilon=8/255$ for the normalized samples. For optimizing PGD perturbations, we applied $r=15$ PGD steps, where we used the standard rule $\alpha=1.5\epsilon/r$ to choose the stepsize parameter $\alpha$. We trained every DNN model for 100 epcohs using the Adam optimizer \cite{kingma2014adam} with a batch-size of $128$. %To avoid label leaking pruth to generaroblem described in \textcolor{red}{PAPER}, we used the prediction of the training model instead of the ground tte adversarial samples. 
The numerical experiments were implemented using the PyTorch \cite{pytorch} platform and were run on one standard RTX-3090 GPU. 
\iffalse
so less $\phi(\mathcal{F})$ indicates better generalization for model $\mathcal{F}$. $\phi(\mathcal{F})$ is formulated as: 
\begin{equation}
    \phi(\mathcal{F})=\mathop{E}_{x\in \mathcal{X}_{Train}}[P(\mathcal{F}(x+p)=y_{gt})]-\mathop{E}_{x\in \mathcal{X}_{test}}[P(\mathcal{F}(x+p)=y_{gt})].
\end{equation}
\fi

In our experiments, we used three standard image recognition datasets: 1) CIFAR-10 \cite{cifar}, 2) CIFAR-100 \cite{cifar}, 3) SVHN \cite{SVHN}, and the following four neural network architectures: 1) AlexNet \cite{AlexNet}, 2) Inception-Net \cite{inceptionv1}, 3) VGG-16 \cite{VGG}, 4) ResNet-18 \cite{ResNet}. In the reported results, we evaluate a prediction model's generalization performance using the accuracy gap between the training and test sets. For evaluating the transferability performance, we used the generated black-box adversarial examples and measured the transferability rate as the target network's averaged classification error over the designed adversarial examples on the test set. Therefore, a higher transferability rate implies more transferable adversarial examples.   

In the numerical evaluation of the adversarial examples' transferability to the target models, we only considered those samples whose original unperturbed versions had been labeled correctly by the target DNN, as we expect the clean version of an adversarial example for the target DNN to be at least labeled correctly by this classifier. %Therefore, we note that the better transferability scores reported for the regularized DNNs is not a mere consequence of their better performance on clean test samples. 
In addition, we used different training sets for the substitute and target classifiers to separate the generalization effects of the substitute and target DNNs. To do this, we split the training set in half and used each half for training one of the two classifiers. Finally, in consistence with our theoretical analysis, we used PGD adversarial training for training the substitute DNN model and applied standard ERM training for training the target DNNs. \vspace{-3mm}

\iffalse
For model $\mathcal{F}$, define the expected transferability of all perturbations generated from $\mathcal{F}$ as $\Omega(\mathcal{F})$, and less $\Omega(\mathcal{F})$ represents better transferability for perturbations generated from $\mathcal{F}$. $\Omega(\mathcal{F})$ can be formulated as:
 \begin{equation}
     \Omega(\mathcal{F})=\mathop{E}_{x\in \mathcal{X}}[P(\mathcal{F}'(x+p)=y_{gt})]
 \end{equation}
 
 It's worth noting that when evaluating $\Omega(\mathcal{F})$, we only generate adversarial perturbations from clean samples when these clean samples can be labeled by $\mathcal{F}$ correctly. This is to make sure that the generated perturbations are representative to model $\mathcal{F}$, and only in this case the transferability of perturbations makes sense.
 \fi

\subsection{Transferability and Generalization under Spectral Regularization}
We evaluated the generalization and transefrability performance of the discussed black-box attack schemes for Lipschitz-regularized neural nets. To apply spectral regularization, we used the spectral normalization method \cite{miyato2018spectral,farnia2018generalizable} constraining the $L_2$-operator norm of the substitute DNN's weight matrices. %As a result, the spectrally-regularized DNN is guaranteed to have a bounded Lipschitz constant scaling with the product of the operator norms. 
We define hyper-parameter $\beta$ as the maximum allowed $L_2$-operator norm. Then, the standard spectral normalization method modifies each weight matrix $W_i$ in \eqref{neural_network_functions} to the following $\widetilde{W}_i$:\vspace{-3mm}
\begin{equation}
    \widetilde{W}_i \, := \, \frac{W_i}{\max\{1,\frac{\Vert W_i\Vert_2}{\beta}\}} \, =\, \begin{cases}
    W_i \quad &\text{\rm if}\; \Vert W_i\Vert_2\le \beta, \\
    \frac{\beta}{\Vert W_i\Vert_2}W_i \quad &\text{\rm otherwise.}
    \end{cases}
\end{equation}
The above operation will regularize the matrix's operator norm to be upper-bounded by $\beta$. %Assuming the spectral norm of all activation functions are upper-bounded by $1$, e.g. the ReLU and ELU activation functions, then the Lipschitz coefficient of the regularized $L$-layer DNN will be bounded by $\beta^L$.

\begin{figure}[htbp]
    \centering
    % \hspace{-0.8cm}\includegraphics[height=4cm]{NeurIPS2021/submission/Figures/Experiment_Fig/tran_L2_FGM_cifar10.png}
    \includegraphics[width=\linewidth]{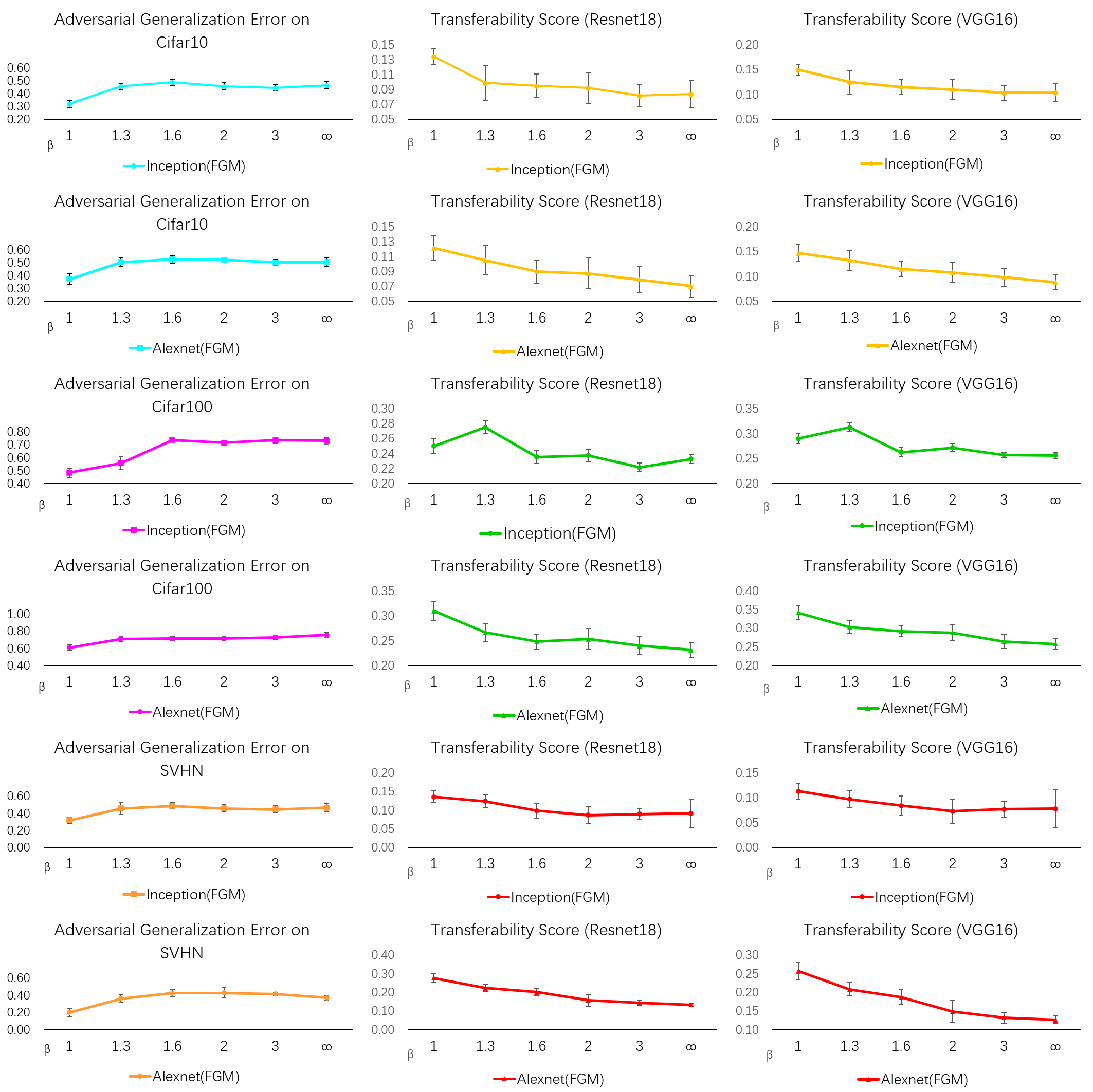}
    \caption{Generalization errors of substitute DNNs (the lower the better), and transferability rates of adversarial examples generated from the substitute model (the higher the better) for CIFAR-10 (rows 1-2), CIFAR-100 (rows 3-4) and SVHN (rows 5-6) datasets. ResNet18 and VGG-16 architectures were used as the target DNNs.}

    % The top row shows the results for AlexNet and Inception trained by FGM adversarial training on CIFAR-10 data. The bottom row shows the results for the Inception-Net trained by $L_\infty$-bounded PGD on SVHN.}
    \label{fig:tran_L2_FGM}
\end{figure}

Figure~\ref{fig:tran_L2_FGM} shows the generalization error of the model and attack transferability rates of the generated perturbations using the substitute classifier AlexNet and Inception-Net under different spectral-norm hyperparameter $\beta$'s.  The numerical results show that in all cases through applying the stronger regularization coefficients $\beta=1.0,1.3$, the AlexNet and Inception classifiers achieve the highest generalization performance and attack transferability rates to the target ResNet-18 and VGG-16. Therefore, spectral regularization not only helped the DNN classifier attain better generalization score, which is an expected outcome, but further improved the transferability of the designed perturbations to unseen DNNs with different architectures. These numerical results suggest the impact of the substitute network's generalization error on the success of the adversarial examples on other DNNs.

\begin{table}[]
\centering
\caption{Generalization error (Gen. Err.) and $L_2$-norm-based adversarial examples' transferability rates on three image datasets, with and without spectral regularization ($\beta=\infty$ means no spectral regularization). %Train($\mathcal{F}$) and Test($\mathcal{F}$) represents the perturbations' transferability, which are generated from training and test samples, respectively. 
%VGG16 and RN18 represent setting target model as VGG16 and ResNet18.
}
\resizebox{0.75\linewidth}{!}
{\begin{tabular}{|ccccccc|}
\hline
Dataset & Model & Method & $\beta$ &Gen.Err. & \begin{tabular}[c]{@{}c@{}}Transferability\\ Rate(VGG16)\end{tabular} & \begin{tabular}[c]{@{}c@{}}Transferability\\ Rate(ResNet18)\end{tabular} \\ \hline
\multicolumn{1}{|c|}{} & \multicolumn{1}{c|}{} & \multicolumn{1}{c|}{} & \multicolumn{1}{c|}{$\infty$} & 0.545 & 0.105 & 0.087 \\ \cline{4-4}
\multicolumn{1}{|c|}{} & \multicolumn{1}{c|}{} & \multicolumn{1}{c|}{\multirow{-2}{*}{PGD}} & \multicolumn{1}{c|}{\cellcolor[HTML]{EFEFEF}1.0} & \cellcolor[HTML]{EFEFEF}0.342 & \cellcolor[HTML]{EFEFEF}0.162 & \cellcolor[HTML]{EFEFEF}0.139 \\ \cline{3-4}
\multicolumn{1}{|c|}{} & \multicolumn{1}{c|}{} & \multicolumn{1}{c|}{} & \multicolumn{1}{c|}{$\infty$} & 0.505 & 0.089 & 0.070 \\ \cline{4-4}
\multicolumn{1}{|c|}{} & \multicolumn{1}{c|}{\multirow{-4}{*}{AlexNet}} & \multicolumn{1}{c|}{\multirow{-2}{*}{FGM}} & \multicolumn{1}{c|}{\cellcolor[HTML]{EFEFEF}1.0} & \cellcolor[HTML]{EFEFEF}0.451 & \cellcolor[HTML]{EFEFEF}0.147 & \cellcolor[HTML]{EFEFEF}0.122 \\ \cline{2-7} 
\multicolumn{1}{|c|}{} & \multicolumn{1}{c|}{} & \multicolumn{1}{c|}{} & \multicolumn{1}{c|}{$\infty$} & 0.508 & 0.104 & 0.084 \\ \cline{4-4}
\multicolumn{1}{|c|}{} & \multicolumn{1}{c|}{} & \multicolumn{1}{c|}{\multirow{-2}{*}{PGD}} & \multicolumn{1}{c|}{\cellcolor[HTML]{EFEFEF}1.0} & \cellcolor[HTML]{EFEFEF}0.258 & \cellcolor[HTML]{EFEFEF}0.150 & \cellcolor[HTML]{EFEFEF}0.134 \\ \cline{3-4}
\multicolumn{1}{|c|}{} & \multicolumn{1}{c|}{} & \multicolumn{1}{c|}{} & \multicolumn{1}{c|}{$\infty$} & 0.466 & 0.092 & 0.078 \\ \cline{4-4}
\multicolumn{1}{|c|}{\multirow{-8}{*}{Cifar10}} & \multicolumn{1}{c|}{\multirow{-4}{*}{Inception}} & \multicolumn{1}{c|}{\multirow{-2}{*}{FGM}} & \multicolumn{1}{c|}{\cellcolor[HTML]{EFEFEF}1.0} & \cellcolor[HTML]{EFEFEF}0.320 & \cellcolor[HTML]{EFEFEF}0.136 & \cellcolor[HTML]{EFEFEF}0.113 \\ \hline
\multicolumn{1}{|c|}{} & \multicolumn{1}{c|}{} & \multicolumn{1}{c|}{} & \multicolumn{1}{c|}{$\infty$} & 0.789 & 0.229 & 0.260 \\ \cline{4-4}
\multicolumn{1}{|c|}{} & \multicolumn{1}{c|}{} & \multicolumn{1}{c|}{\multirow{-2}{*}{PGD}} & \multicolumn{1}{c|}{\cellcolor[HTML]{EFEFEF}1.0} & \cellcolor[HTML]{EFEFEF}0.601 & \cellcolor[HTML]{EFEFEF}0.323 & \cellcolor[HTML]{EFEFEF}0.353 \\ \cline{3-4}
\multicolumn{1}{|c|}{} & \multicolumn{1}{c|}{} & \multicolumn{1}{c|}{} & \multicolumn{1}{c|}{$\infty$} & 0.758 & 0.258 & 0.232 \\ \cline{4-4}
\multicolumn{1}{|c|}{} & \multicolumn{1}{c|}{\multirow{-4}{*}{AlexNet}} & \multicolumn{1}{c|}{\multirow{-2}{*}{FGM}} & \multicolumn{1}{c|}{\cellcolor[HTML]{EFEFEF}1.0} & \cellcolor[HTML]{EFEFEF}0.611 & \cellcolor[HTML]{EFEFEF}0.342 & \cellcolor[HTML]{EFEFEF}0.310 \\ \cline{2-7} 
\multicolumn{1}{|c|}{} & \multicolumn{1}{c|}{} & \multicolumn{1}{c|}{} & \multicolumn{1}{c|}{$\infty$} & 0.602 & 0.303 & 0.270 \\ \cline{4-4}
\multicolumn{1}{|c|}{} & \multicolumn{1}{c|}{} & \multicolumn{1}{c|}{\multirow{-2}{*}{PGD}} & \multicolumn{1}{c|}{\cellcolor[HTML]{EFEFEF}1.3} & \cellcolor[HTML]{EFEFEF}0.494 & \cellcolor[HTML]{EFEFEF}0.330 & \cellcolor[HTML]{EFEFEF}0.301 \\ \cline{3-4}
\multicolumn{1}{|c|}{} & \multicolumn{1}{c|}{} & \multicolumn{1}{c|}{} & \multicolumn{1}{c|}{$\infty$} & 0.717 & 0.268 & 0.236 \\ \cline{4-4}
\multicolumn{1}{|c|}{\multirow{-8}{*}{Cifar100}} & \multicolumn{1}{c|}{\multirow{-4}{*}{Inception}} & \multicolumn{1}{c|}{\multirow{-2}{*}{FGM}} & \multicolumn{1}{c|}{\cellcolor[HTML]{EFEFEF}1.3} & \cellcolor[HTML]{EFEFEF}0.558 & \cellcolor[HTML]{EFEFEF}0.313 & \cellcolor[HTML]{EFEFEF}0.275 \\ \hline
\multicolumn{1}{|c|}{} & \multicolumn{1}{c|}{} & \multicolumn{1}{c|}{} & \multicolumn{1}{c|}{$\infty$} & 0.298 & 0.211 & 0.225 \\ \cline{4-4}
\multicolumn{1}{|c|}{} & \multicolumn{1}{c|}{} & \multicolumn{1}{c|}{\multirow{-2}{*}{PGD}} & \multicolumn{1}{c|}{\cellcolor[HTML]{EFEFEF}1.0} & \cellcolor[HTML]{EFEFEF}0.199 & \cellcolor[HTML]{EFEFEF}0.276 & \cellcolor[HTML]{EFEFEF}0.292 \\ \cline{3-4}
\multicolumn{1}{|c|}{} & \multicolumn{1}{c|}{} & \multicolumn{1}{c|}{} & \multicolumn{1}{c|}{$\infty$} & 0.373 & 0.134 & 0.126 \\ \cline{4-4}
\multicolumn{1}{|c|}{} & \multicolumn{1}{c|}{\multirow{-4}{*}{AlexNet}} & \multicolumn{1}{c|}{\multirow{-2}{*}{FGM}} & \multicolumn{1}{c|}{\cellcolor[HTML]{EFEFEF}1.0} & \cellcolor[HTML]{EFEFEF}0.203 & \cellcolor[HTML]{EFEFEF}0.277 & \cellcolor[HTML]{EFEFEF}0.257 \\ \cline{2-7} 
\multicolumn{1}{|c|}{} & \multicolumn{1}{c|}{} & \multicolumn{1}{c|}{} & \multicolumn{1}{c|}{$\infty$} & 0.342 & 0.193 & 0.177 \\ \cline{4-4}
\multicolumn{1}{|c|}{} & \multicolumn{1}{c|}{} & \multicolumn{1}{c|}{\multirow{-2}{*}{PGD}} & \multicolumn{1}{c|}{\cellcolor[HTML]{EFEFEF}1.0} & \cellcolor[HTML]{EFEFEF}0.115 & \cellcolor[HTML]{EFEFEF}0.339 & \cellcolor[HTML]{EFEFEF}0.313 \\ \cline{3-4}
\multicolumn{1}{|c|}{} & \multicolumn{1}{c|}{} & \multicolumn{1}{c|}{} & \multicolumn{1}{c|}{$\infty$} & 0.373 & 0.134 & 0.126 \\ \cline{4-4}
\multicolumn{1}{|c|}{\multirow{-8}{*}{SVHN}} & \multicolumn{1}{c|}{\multirow{-4}{*}{Inception}} & \multicolumn{1}{c|}{\multirow{-2}{*}{FGM}} & \multicolumn{1}{c|}{\cellcolor[HTML]{EFEFEF}1.0} & \cellcolor[HTML]{EFEFEF}0.203 & \cellcolor[HTML]{EFEFEF}0.277 & \cellcolor[HTML]{EFEFEF}0.257 \\ \hline
\end{tabular}}
\label{tab:generalization_transferability_SN}
\end{table}

Table~\ref{tab:generalization_transferability_SN} shows our numerical results validating the connection between the substitute DNN's generalization and $L_2$-norm-based designed adversarial examples' transferability. In this table, we report the performance of spectral regularization under the best $\beta$ hyperparameter for validation samples. As can be seen in this table, spectral regularization manages to consistently improve the transferability rates of the adversarial examples, which confirms our hypothesis that better generalization will lead to more transferable adversarial examples. Due to the 9-page space limit, we defer the presentation of our numerical results for $L_\infty$-norm-based adversarial examples to the Appendix.\vspace{-3mm}

%Surprisingly, in all these experiments, \textcolor{red}{There is a strict positive correlation between generalization and migration}.

\subsection{Transferability and Generalization under Early Stopping}
Next, we used the implicit regularization mechanism of early stopping \cite{yao2007early} to validate that better generalization achieved under early stopping can help generating more transferable adversarial examples. To perform early stopping, we used 30\% of the original test set as the validation set, and used the remaining 70\% to measure the test accuracy. We stopped the DNN training when the trained model achieved its best performance on the validation samples. Due to the 9-page space limit, here we present the CIFAR-10 and SVHN numerical results in Table~\ref{tab:early_stopping}. The complete set of obtained numerical results is deferred to the Appendix. Again, our numerical results suggest that both the generalization and transferability scores considerably improve under the early stopping regularization. The observation is consistent with our hypothesis on the impact of the generalization behavior of the substitute network on the transferability of adversarial examples.

\begin{table}
\centering
\caption{Generalization error (Gen. Err.) and adversarial examples' transferability rates on three image datasets, with and without early stopping (ES)}
\resizebox{0.75\linewidth}{!}{\begin{tabular}{|cccccc|}
\hline
Dataset & Model & Method & Gen. Err. & \begin{tabular}[c]{@{}c@{}}Transferability\\ Rate(VGG16)\end{tabular} & \begin{tabular}[c]{@{}c@{}}Transferability\\ Rate(ResNet18)\end{tabular} \\ \hline
\multicolumn{1}{|c|}{} & \multicolumn{1}{c|}{} & \multicolumn{1}{c|}{PGD} & 0.517 & 0.127 & 0.104 \\
\multicolumn{1}{|c|}{} & \multicolumn{1}{c|}{} & \multicolumn{1}{c|}{\cellcolor[HTML]{EFEFEF}PGD-ES} & \cellcolor[HTML]{EFEFEF}0.073 & \cellcolor[HTML]{EFEFEF}0.198 & \cellcolor[HTML]{EFEFEF}0.172 \\ \cline{3-3}
\multicolumn{1}{|c|}{} & \multicolumn{1}{c|}{} & \multicolumn{1}{c|}{FGM} & 0.467 & 0.100 & 0.089 \\
\multicolumn{1}{|c|}{} & \multicolumn{1}{c|}{\multirow{-4}{*}{Inception}} & \multicolumn{1}{c|}{\cellcolor[HTML]{EFEFEF}FGM-ES} & \cellcolor[HTML]{EFEFEF}0.126 & \cellcolor[HTML]{EFEFEF}0.170 & \cellcolor[HTML]{EFEFEF}0.147 \\ \cline{2-6} 
\multicolumn{1}{|c|}{} & \multicolumn{1}{c|}{} & \multicolumn{1}{c|}{PGD} & 0.579 & 0.098 & 0.077 \\
\multicolumn{1}{|c|}{} & \multicolumn{1}{c|}{} & \multicolumn{1}{c|}{\cellcolor[HTML]{EFEFEF}PGD-ES} & \cellcolor[HTML]{EFEFEF}0.061 & \cellcolor[HTML]{EFEFEF}0.154 & \cellcolor[HTML]{EFEFEF}0.136 \\ \cline{3-3}
\multicolumn{1}{|c|}{} & \multicolumn{1}{c|}{} & \multicolumn{1}{c|}{FGM} & 0.520 & 0.100 & 0.087 \\
\multicolumn{1}{|c|}{\multirow{-8}{*}{Cifar10}} & \multicolumn{1}{c|}{\multirow{-4}{*}{AlexNet}} & \multicolumn{1}{c|}{\cellcolor[HTML]{EFEFEF}FGM-ES} & \cellcolor[HTML]{EFEFEF}0.092 & \cellcolor[HTML]{EFEFEF}0.152 & \cellcolor[HTML]{EFEFEF}0.127 \\ \hline
\multicolumn{1}{|c|}{} & \multicolumn{1}{c|}{} & \multicolumn{1}{c|}{PGD} & 0.341 & 0.207 & 0.220 \\
\multicolumn{1}{|c|}{} & \multicolumn{1}{c|}{} & \multicolumn{1}{c|}{\cellcolor[HTML]{EFEFEF}PGD-ES} & \cellcolor[HTML]{EFEFEF}0.057 & \cellcolor[HTML]{EFEFEF}0.298 & \cellcolor[HTML]{EFEFEF}0.322 \\ \cline{3-3}
\multicolumn{1}{|c|}{} & \multicolumn{1}{c|}{} & \multicolumn{1}{c|}{FGM} & 0.380 & 0.136 & 0.129 \\
\multicolumn{1}{|c|}{} & \multicolumn{1}{c|}{\multirow{-4}{*}{Inception}} & \multicolumn{1}{c|}{\cellcolor[HTML]{EFEFEF}FGM-ES} & \cellcolor[HTML]{EFEFEF}0.180 & \cellcolor[HTML]{EFEFEF}0.213 & \cellcolor[HTML]{EFEFEF}0.219 \\ \cline{2-6} 
\multicolumn{1}{|c|}{} & \multicolumn{1}{c|}{} & \multicolumn{1}{c|}{PGD} & 0.307 & 0.211 & 0.228 \\
\multicolumn{1}{|c|}{} & \multicolumn{1}{c|}{} & \multicolumn{1}{c|}{\cellcolor[HTML]{EFEFEF}PGD-ES} & \cellcolor[HTML]{EFEFEF}0.030 & \cellcolor[HTML]{EFEFEF}0.256 & \cellcolor[HTML]{EFEFEF}0.278 \\ \cline{3-3}
\multicolumn{1}{|c|}{} & \multicolumn{1}{c|}{} & \multicolumn{1}{c|}{FGM} & 0.373 & 0.157 & 0.170 \\
\multicolumn{1}{|c|}{\multirow{-8}{*}{SVHN}} & \multicolumn{1}{c|}{\multirow{-4}{*}{AlexNet}} & \multicolumn{1}{c|}{\cellcolor[HTML]{EFEFEF}FGM-ES} & \cellcolor[HTML]{EFEFEF}0.064 & \cellcolor[HTML]{EFEFEF}0.241 & \cellcolor[HTML]{EFEFEF}0.260 \\ \hline
\end{tabular}}
\label{tab:early_stopping}
\end{table}

\begin{figure}
    \centering
    \includegraphics[width=\linewidth]{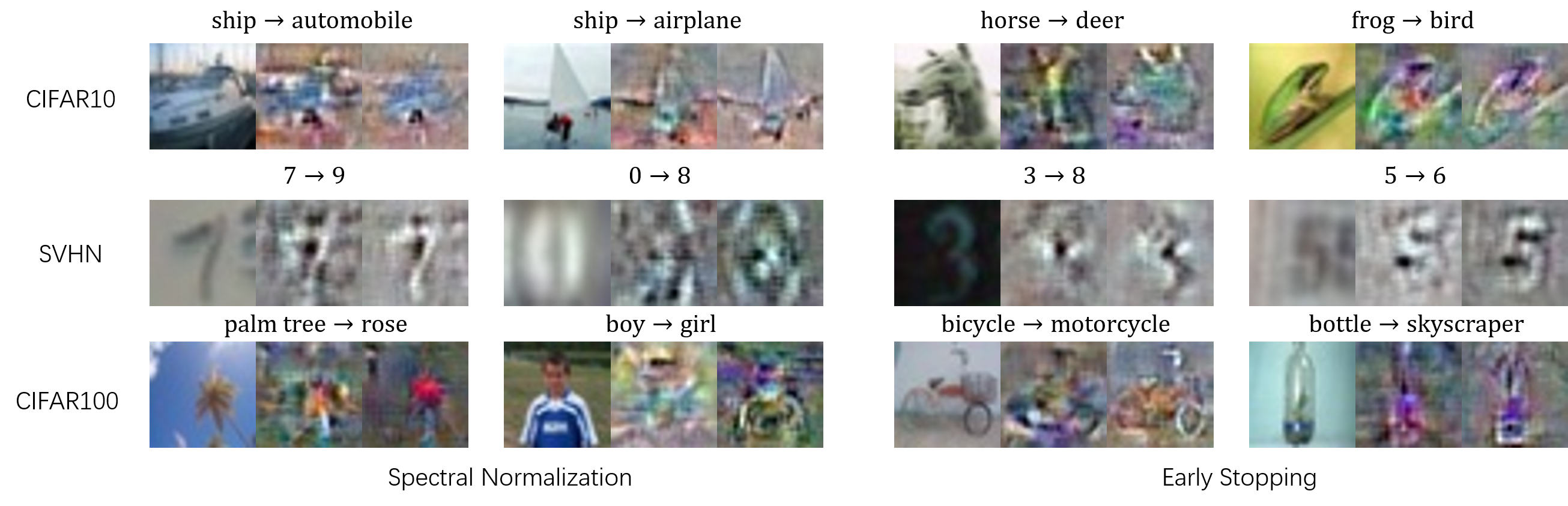}
    \caption{Visualization of adversarial perturbations. Each set of three pictures shows the original sample, the untransferable perturbation from the unregularized DNN, and the transferable perturbation generated by the regularized model (left to right). $A\rightarrow B$ indicates the groundtruth label $A$ and the transferable example's predicted label $B$.%. When the intransferable perturbation can not fool the target model, the transferable perturbation can fool the target model to predict it as $B$.
    }
    \label{fig:visualization}
\end{figure}

Finally, Figure~\ref{fig:visualization} illustrates 12 uniformly-sampled transferable adversarial examples under spectral regularization and early stopping. We note that the adversarial examples designed by the unregularized DNN for these test samples failed to transfer to the target DNNs. We also observed that the transferable perturbations generated from a regularized DNN had sharper edges and less noise power in the background, and concentrated the perturbation's power on the central part of the image. 
%These examples showcase the effect of the substitute DNN's generalization behavior on the success of the designed adversarial examples. 

% remove training transferability

% no reg
% training with different number pf training samples

% verify 19.8 in 45.3

% use intersection

% \vspace{-.14cm}
\section{Conclusion}
In this paper, we provided theoretical and numerical evidence on how the generalization properties of a substitute neural network can influence the transferability of the generated adversarial examples to other classifiers. While the transferability of black-box adversarial attacks and generalization power of the substitute classifier may seem two orthogonal factors, our results indicate existing interconnections between the two aspects. We note that our work develops standard uniform-convergence generalization bounds, and an interesting future direction is to apply the recently-developed generalization analysis of overparameterized function spaces to gain a better understanding of the role of benign overfitting in the transferability of adversarial examples. Also, our experimental results motivate further studies of how other popular regularization methods in deep learning, such as batch normalization and dropout, can affect the transferability of designed adversarial examples.

\bibliographystyle{unsrt}  
\bibliography{ref}

%\newpage
%\section*{Appendix}
%\appendix
% \begin{center}
%     \large{Supplementary material for paper: \\ \vspace{.2cm}
%     \textbf{On the Role of Generalization in Transferability of
% Adversarial Perturbations}}
% \end{center}

% \vspace{.7cm}

\begin{appendices}
\section{Proofs}
\subsection{Proof of Proposition \ref{Prop: correspondence}}
To prove the proposition, note that the optimization problem for $\lambda$-optimal adversarial attack scheme can be written as
\begin{equation*}
    \max_{\delta:\mathcal{X}\times \mathcal{Y}\rightarrow \mathbb{R}^d}\; \mathbb{E}\biggl[\ell\bigl(h_\mathbf{w}(\mathbf{X}+\delta(\mathbf{X},Y)),Y\bigr) - \frac{\lambda}{2}\Vert  \delta(\mathbf{X},Y) \Vert^2\biggr].
\end{equation*}
We observe that the above optimization problem decouples into separate problems for every $(\mathbf{x},y)$, and hence $\delta^*(\mathbf{x},y)$ is the optimal solution to the following optimization problem
\begin{equation*}
    \max_{\boldsymbol{\delta}\in\mathbb{R}^d}\; \ell\bigl(h_\mathbf{w}(\mathbf{x}+\boldsymbol{\delta}),y\bigr) - \frac{\lambda}{2}\Vert  \boldsymbol{\delta} \Vert^2.
\end{equation*}
Since $\ell \circ h_{\mathbf{w}}$ is assumed to be $\lambda$-smooth in $\mathbf{x}$, the objective function in the above optimization problem is a concave function of $\boldsymbol{\delta}$. This is because the Hessian of the above objective function will be negative semi-definite as
\begin{equation}
     \nabla_{\boldsymbol{\delta}}^2 \biggl[\ell\bigl(h_\mathbf{w}(\mathbf{x}+\boldsymbol{\delta}),y\bigr) - \frac{\lambda}{2}\Vert  \boldsymbol{\delta} \Vert^2 \biggr] =  \nabla_{\boldsymbol{\delta}}^2 \ell\bigl(h_\mathbf{w}(\mathbf{x}+\boldsymbol{\delta}),y\bigr) -\lambda I_{d\times d} \preceq \mathbf{0}.
\end{equation}
Therefore, applying the first-order necessary condition implies that a globally optimal solution $\delta^*(\mathbf{x},y)$ to the above concave objective function will be the solution to
\begin{equation*}
    \nabla_{\mathbf{x}}\ell\bigl(h_\mathbf{w}(\mathbf{x}+\delta^*(\mathbf{x},y)),y\bigr) - \lambda \delta^*(\mathbf{x},y) = \mathbf{0}. 
\end{equation*}
The above necessary and sufficient condition for $\delta^*(\mathbf{x},y)$ can be rewritten as:
\begin{equation*}
    \bigl( \mathbf{x} + \delta^*(\mathbf{x},y)\bigr) -\frac{1}{\lambda}\nabla_{\mathbf{x}}\ell\bigl(h_\mathbf{w}(\mathbf{x}+\delta^*(\mathbf{x},y)),y\bigr) = \mathbf{x}. 
\end{equation*}
Note that this condition is equivalent to the following equation which completes the proof:
\begin{equation*}
    \delta^*(\mathbf{x},y) \, =\,   \left(\text{\rm Id}_{\mathbf{x}} - \frac{1}{\lambda}\nabla_{\mathbf{x}}\ell\circ h_\mathbf{w}\right)^{-1}(\mathbf{x}) \, - \, \mathbf{x}.
\end{equation*}

\subsection{Proof of Theorem \ref{Thm: generalization bound}}
To show Theorem \ref{Thm: generalization bound}, we first present the following lemmas.
\begin{lemma}[\cite{farnia2018generalizable}, Lemma 7]\label{Lemma: smooth DNN}
Under Assumptions \ref{Assumption: loss}, \ref{Assumption: substitute neural network}, the substitute neural network's loss function $\ell(h_{\mathbf{w}}(\mathbf{x}),y)$ is $\kappa$-smooth in input vector $\mathbf{x}$, i.e its gradient with respect to $\mathbf{x}$ is $\kappa$-Lipschitz and satisfies
\begin{equation*}
    \forall \mathbf{x},\mathbf{x}'\in\mathcal{X},\, y\in\mathcal{Y}:\quad \big\Vert \nabla_{\mathbf{x}}\ell\bigl(h_{\mathbf{w}}(\mathbf{x}),y\bigr) - \nabla_{\mathbf{x}}\ell\bigl(h_{\mathbf{w}}(\mathbf{x}'),y\bigr)\big\Vert \le \kappa \Vert \mathbf{x} - \mathbf{x}'\Vert,
\end{equation*}
where $\kappa = \left(\sum_{i=0}^L \prod_{j=0}^i \gamma_j\Vert W_j\Vert\right)\left(\prod_{i=0}^L \gamma_i\Vert W_i\Vert\right)$.
\end{lemma}

\begin{lemma}
Under Theorem \ref{Thm: generalization bound}'s assumptions, the $\lambda$-optimal attack scheme $\delta^*_\mathbf{w}:\mathcal{X}\times\mathcal{Y}\rightarrow \mathbb{R}^d$ satisfies the following output norm constraint for every $(\mathbf{x},y)\in \mathcal{X}\times\mathcal{Y}$:
\begin{equation*}
    \big\Vert \delta^*_\mathbf{w}(\mathbf{x},y)\big\Vert \le \frac{\prod_{i=0}^L \gamma_i\Vert W_i\Vert_2}{\lambda} = \frac{L_{\mathbf{w}}}{\lambda}.
\end{equation*}
\end{lemma}
\begin{proof}
Note that since $\lambda > \left(\sum_{i=0}^L \prod_{j=0}^i \gamma_j\Vert W_j\Vert_2\right)\prod_{i=0}^L \gamma_i\Vert W_i\Vert_2$ holds according to Theorem \ref{Thm: generalization bound}'s assumptions, the smoothness condition of Proposition \ref{Prop: correspondence} will hold according to Lemma \ref{Lemma: smooth DNN}. As a result, we have
\begin{equation*}
    {\delta}^*_{\mathbf{w}}(\mathbf{x},y) = \frac{1}{\lambda}\nabla_{\mathbf{x}}\ell\bigl(h_{\mathbf{w}}(\mathbf{x}+{\delta}^*_{\mathbf{w}}(\mathbf{x},y)),y\bigr).
\end{equation*}
Therefore,
\begin{equation*}
    \big\Vert{\delta}^*_{\mathbf{w}}(\mathbf{x},y)\big\Vert \le \frac{\Vert \nabla_{\mathbf{x}}\ell\bigl(h_{\mathbf{w}}(\mathbf{x}+{\delta}^*_{\mathbf{w}}(\mathbf{x},y)),y\bigr)\Vert}{\lambda} \le \frac{\prod_{i=0}^L \gamma_i\Vert W_i\Vert_2}{\lambda}.
\end{equation*}
The final inequality follows from the Lipschitz coefficient of DNN function $h_{\mathbf{w}}$ which is the composition of linear transformation $W_i$'s (with Lipschitz constant $\Vert W_i\Vert_2$) and activation non-linearity $\phi_i$'s  (with Lipschitz constant $\gamma_i$). Therefore, the proof is complete.
\end{proof}

\begin{lemma}\label{Lemma: Perturbation FGM}
Under Assumption \ref{Assumption: substitute neural network}, the substitute neural network $h_{\mathbf{w}}$'s gradient satisfies the following error bound under a perturbation  matrix $\Delta_k$ with $L_2$-operator norm $\Vert\Delta_k\Vert_2\le t$ to wight matrix $W_k$, where we define $\widetilde{\mathbf{w}}=\operatorname{vec}(W_0,\ldots,W_{k-1},W_k+\Delta_k,W_{k+1},\ldots,W_L)$:
\begin{equation*}
    \big\Vert \nabla_{\mathbf{x}}h_{\mathbf{w}}(\mathbf{x}) - \nabla_{\mathbf{x}}h_{\widetilde{\mathbf{w}}}(\mathbf{x})\big\Vert \le \frac{L_{\mathbf{w}}\sum_{i=k}^L \prod_{j=0}^i \gamma_j\Vert W_j\Vert}{\Vert W_k\Vert_2}{\Vert\Delta_k\Vert_2}
\end{equation*}
\end{lemma}
\begin{proof}
Note that the neural network's Jacobian with respect to the input follows from:
\begin{align*}
    \mathrm{J}_{h_{\mathbf{w}}}(\mathbf{x}) = \prod_{i=0}^L W_i^\top \operatorname{diag}\bigl( \phi'_i(h_{\mathbf{w_{0:i}}}(\mathbf{x})) \bigr).
\end{align*}
In the above, $h_{\mathbf{w_{0:i}}}(\mathbf{x})$ denotes the neural net's output at layer $i$. Therefore, for $\widetilde{\mathbf{w}}$ which is different from $\mathbf{w}$ only at layer $k$ we will have:
\begin{align*}
    \big\Vert \mathrm{J}_{h_{\mathbf{w}}}(\mathbf{x}) - \mathrm{J}_{h_{\widetilde{\mathbf{w}}}}(\mathbf{x}) \big\Vert_2 \, &\le \, \sum_{i=k}^L\biggl[ \bigl(\prod_{j=0}^L \gamma_j\Vert W_j \Vert_2\bigr) \bigl(\prod_{j=0}^i \gamma_j\Vert W_j \Vert_2\bigr)\biggr] \frac{\Vert\Delta_k\Vert_2}{\Vert W_k\Vert_2} \\
    &= \, \biggl(\prod_{j=0}^L \gamma_j\Vert W_j \Vert_2\biggr)\sum_{i=k}^L\biggl[  \prod_{j=0}^i \gamma_j\Vert W_j \Vert_2\biggr] \frac{\Vert\Delta_k\Vert_2}{\Vert W_k\Vert_2} \\
    &= \, \frac{L_{\mathbf{w}}\sum_{i=k}^L \prod_{j=0}^i \gamma_j\Vert W_j\Vert}{\Vert W_k\Vert_2}{\Vert\Delta_k\Vert_2}.
\end{align*}
The proof is hence finished.
\end{proof}

\begin{lemma}\label{Lemma: Perturbation WRM}
Under Theorem \ref{Thm: generalization bound}'s assumptions, the $\lambda$-optimal attack scheme $\delta^*_\mathbf{w}:\mathcal{X}\times\mathcal{Y}\rightarrow \mathbb{R}^d$ satisfies the following error bound under a norm-bounded perturbation $\Delta_k:\: \Vert\Delta_k\Vert_2\le t$: to wight matrix $W_k$ where we define $\widetilde{\mathbf{w}}=\operatorname{vec}(W_0,\ldots,W_{k-1},W_k+\Delta_j,W_{k+1},\ldots,W_L)$:
\begin{equation*}
    \big\Vert \delta^*_{\mathbf{w}}(\mathbf{x},y) - \delta^*_{\widetilde{\mathbf{w}}}(\mathbf{x},y)\big\Vert \le \frac{L_{\mathbf{w}}\sum_{i=k}^L \prod_{j=0}^i \gamma_j\Vert W_j\Vert}{\tau\lambda \Vert W_k\Vert_2 }\Vert\Delta_k\Vert_2
\end{equation*}
\end{lemma}
\begin{proof}
Since $\lambda > \left(\sum_{i=0}^L \prod_{j=0}^i \gamma_j\Vert W_j\Vert_2\right)\prod_{i=0}^L \gamma_i\Vert W_i\Vert_2$ follows from Theorem \ref{Thm: generalization bound}'s assumption, Proposition \ref{Prop: correspondence} will hold according to Lemma \ref{Lemma: smooth DNN} and implies that
\begin{equation*}
    {\delta}^*_{\mathbf{w}}(\mathbf{x},y) = \frac{1}{\lambda}\nabla_{\mathbf{x}}\ell\bigl(h_{\mathbf{w}}(\mathbf{x}+{\delta}^*_{\mathbf{w}}(\mathbf{x},y)),y\bigr).
\end{equation*}
As a result,
\begin{align*}
    &\big\Vert {\delta}^*_{\mathbf{w}}(\mathbf{x},y) - {\delta}^*_{\widetilde{\mathbf{w}}}(\mathbf{x},y)\big\Vert \\
    = \, & \frac{1}{\lambda}\biggl\Vert \nabla_{\mathbf{x}}\ell\bigl(h_{\mathbf{w}}(\mathbf{x}+{\delta}^*_{\mathbf{w}}(\mathbf{x},y)),y\bigr) - \nabla_{\mathbf{x}}\ell\bigl(h_{\widetilde{\mathbf{w}}}(\mathbf{x}+{\delta}^*_{\widetilde{\mathbf{w}}}(\mathbf{x},y)),y\bigr)\biggr\Vert \\
    {\le} \, & \frac{1}{\lambda}\biggl\Vert \nabla_{\mathbf{x}}\ell\bigl(h_{\mathbf{w}}(\mathbf{x}+{\delta}^*_{\mathbf{w}}(\mathbf{x},y)),y\bigr) - \nabla_{\mathbf{x}}\ell\bigl(h_{\mathbf{w}}(\mathbf{x}+{\delta}^*_{\widetilde{\mathbf{w}}}(\mathbf{x},y)),y\bigr)\biggr\Vert \, \\
    &\quad + \, \frac{1}{\lambda}\biggl\Vert \nabla_{\mathbf{x}}\ell\bigl(h_{\mathbf{w}}(\mathbf{x}+{\delta}^*_{\widetilde{\mathbf{w}}}(\mathbf{x},y)),y\bigr) - \nabla_{\mathbf{x}}\ell\bigl(h_{\widetilde{\mathbf{w}}}(\mathbf{x}+{\delta}^*_{\widetilde{\mathbf{w}}}(\mathbf{x},y)),y\bigr)\biggr\Vert \\
     \stackrel{(a)}{\le}  \, & \frac{\operatorname{Lip}(\nabla_{\mathbf{x}}\ell \circ h_{\mathbf{w}} )}{\lambda} \big\Vert {\delta}^*_{\mathbf{w}}(\mathbf{x},y) - {\delta}^*_{\widetilde{\mathbf{w}}}(\mathbf{x},y)\big\Vert \, + \, \frac{L_{\mathbf{w}}\sum_{i=k}^L \prod_{j=0}^i \gamma_j\Vert W_j\Vert}{\lambda\Vert W_k\Vert_2 }{\Vert\Delta_k\Vert_2} \\
     \stackrel{(b)}{\le}  \, & \frac{L_{\mathbf{w}}\sum_{i=k}^L \prod_{j=0}^i \gamma_j\Vert W_j\Vert}{\lambda} \big\Vert {\delta}^*_{\mathbf{w}}(\mathbf{x},y) - {\delta}^*_{\widetilde{\mathbf{w}}}(\mathbf{x},y)\big\Vert \, + \, \frac{L_{\mathbf{w}}\sum_{i=0}^L \prod_{j=0}^i \gamma_j\Vert W_j\Vert}{\lambda\Vert W_k\Vert_2}{\Vert\Delta_k\Vert_2} \\
     \stackrel{(c)}{\le}  \, & (1-\tau) \big\Vert {\delta}^*_{\mathbf{w}}(\mathbf{x},y) - {\delta}^*_{\widetilde{\mathbf{w}}}(\mathbf{x},y)\big\Vert \, + \, \frac{L_{\mathbf{w}}\sum_{i=k}^L \prod_{j=0}^i \gamma_j\Vert W_j\Vert}{\lambda\Vert W_k\Vert_2}{\Vert\Delta_k\Vert_2}
\end{align*}
Here, $(a)$ follows from the definition of Lipschitz constant and Lemma \ref{Lemma: Perturbation FGM}'s weight perturbation bound. $(b)$ comes from a direct application of Lemma \ref{Lemma: smooth DNN}, and $(c)$ follows from Theorem \ref{Thm: generalization bound}' assumption. Therefore, the above inequalities collectively lead to the following bound, which completes the proof:
\begin{equation*}
   \tau\,\big\Vert {\delta}^*_{\mathbf{w}}(\mathbf{x},y) - {\delta}^*_{\widetilde{\mathbf{w}}}(\mathbf{x},y)\big\Vert \; \le \; \frac{L_{\mathbf{w}}\sum_{i=k}^L \prod_{j=0}^i \gamma_j\Vert W_j\Vert}{\lambda\Vert W_k\Vert_2}{\Vert\Delta_k\Vert_2}.
\end{equation*}
\end{proof}

To prove Theorem \ref{Thm: generalization bound}, we follow a covering-number-based approach similar to the generalization analysis in \cite{bartlett2017spectrally} for the standard non-adversarial deep supervised learning problem. To do this, we consider the norm constraints $\Vert W_i\Vert_2\le a'_i,\, \Vert W_i\Vert_{2,1}\le b'_i$ for every $i=0,\ldots,L$, and $\Vert V_j\Vert_2\le a_i,\, \Vert V_j\Vert_{2,1}\le b_i$ for every $j=0,\ldots,K$. Now, we define the following covering resolution parameters for the classifier and substitute DNNs' different layers:
\begin{align*}
    \epsilon'_k = \frac{{\tau\lambda a'_k }\alpha'_k \epsilon}{2(\prod_{i=0}^K \xi_i a_i)(\prod_{i=0}^L \gamma_i a'_i)\bigl(\sum_{i=k}^L \prod_{j=0}^i \gamma_j a'_j)}, \; &\text{\rm where }\; \alpha'_k =\frac{1}{A'}\frac{{b'_k}^{2/3}}{{a'_k}^{2/3}} ,\; A'=\sum_{i=0}^L \frac{{b'_i}^{2/3}}{{a'_i}^{2/3}} \\
    \epsilon_j = \frac{{a_j }\alpha_j \epsilon}{2\prod_{i=j}^K \gamma_i a_i}, \; &\text{\rm where }\; \alpha_j =\frac{1}{A}\frac{b_j^{2/3}}{a_j^{2/3}} ,\; A=\sum_{i=0}^K \frac{b_i^{2/3}}{a_i^{2/3}}.
\end{align*}
Note that Lemma \ref{Lemma: Perturbation WRM} implies that by finding an $\epsilon'_k$-covering for each $W_k$ and $\epsilon_j$-covering for each $V_j$, the covering resolution for $\mathcal{F}\circ\Delta_H|_S$ will be upper-bounded by
\begin{equation*}
    \sum_{k=0}^L\biggl[ \frac{L_{\mathbf{w}}L_{\mathbf{v}}\sum_{i=k}^L \prod_{j=0}^i \gamma_j\Vert W_j\Vert}{{\tau\lambda \Vert W_k\Vert_2 }}\epsilon'_k\biggr] + \sum_{k=0}^K\biggl[\frac{\prod_{i=k}^K \gamma_i\Vert V_i\Vert_2}{{\Vert V_k\Vert_2 }}\epsilon_k\biggr] = \epsilon.
\end{equation*}
Therefore, by applying Lemma A.7 from \cite{bartlett2017spectrally} we will have the following bound on the $\epsilon$-covering-number for the set $\mathcal{F}\circ\Delta_H|_S = \{\ell(f_{\mathbf{v}}(\mathbf{X}+\delta^*_{\mathbf{w}}({\mathbf{X}},Y) ) :\; \forall 0\le i\le K: \Vert V_i\Vert_2\le a_i,\, \Vert V_i\Vert_{2,1}\le b_i,\; \forall 0\le j\le L:\: \Vert W_j\Vert_2\le a'_j,\, \Vert W_j\Vert_{2,1}\le b'_j   \}$

\begin{align*}
    &\log \mathcal{N}\bigl(\mathcal{F}\circ\Delta_H|_S , \Vert\cdot\Vert_2,\epsilon \bigr) \, \\
    \le \, &\sum_{i=0}^L \sup_{{\mathbf{w}}_{-i}, \mathbf{v}\in\mathcal{W,V}}\biggl[ \log \mathcal{N}\bigl(\bigl\{\delta^*_{\mathbf{w}}(\mathbf{X},Y):\: \Vert \mathbf{W}_i\Vert_2\le a'_i ,\, \Vert \mathbf{W}_i\Vert_{2,1}\le b'_i \bigr\}, \Vert\cdot\Vert_2,\epsilon'_i \bigr) \biggr] \\
    &\; + \sum_{i=0}^K \sup_{{\mathbf{w}}, \mathbf{v}_{-i}\in\mathcal{W,V}}\biggl[ \log \mathcal{N}\bigl(\bigl\{ h_{\mathbf{v}_{0:i}}(\mathbf{X}),Y):\: \Vert \mathbf{V}_i\Vert_2\le a_i ,\, \Vert \mathbf{V}_i\Vert_{2,1}\le b_i \bigr\}, \Vert\cdot\Vert_2,\epsilon_i \bigr) \biggr] \\
    \le \, &\sum_{i=0}^L \sup_{{\mathbf{w}}_{-i}, \mathbf{v}\in\mathcal{W,V}}\biggl[ \log \mathcal{N}\bigl(\bigl\{\delta^*_{\mathbf{w}}(\mathbf{X},Y):\:  \Vert \mathbf{W}_i\Vert_{2,1}\le b'_i \bigr\}, \Vert\cdot\Vert_2,\epsilon'_i \bigr) \biggr] \\
    &\; + \sum_{i=0}^K \sup_{{\mathbf{w}}, \mathbf{v}_{-i}\in\mathcal{W,V}}\biggl[ \log \mathcal{N}\bigl(\bigl\{ h_{\mathbf{v}_{0:i}}(\mathbf{X}),Y):\:  \Vert \mathbf{V}_i\Vert_{2,1}\le b_i \bigr\}, \Vert\cdot\Vert_2,\epsilon_i \bigr) \biggr] \\
    \le \, &\sum_{i=0}^L \biggl[ \sup_{{\mathbf{w}}_{-i}, \mathbf{v}\in\mathcal{W,V}} \frac{{b'_i}^2\Vert \delta^*_{\mathbf{w}}(\mathbf{X},Y)\Vert_2^2}{{\epsilon'_i}^2} \log(2W^2) \biggr] \\
    &\; + \sum_{i=0}^L \biggl[ \sup_{{\mathbf{w}}, \mathbf{v}_{-i}\in\mathcal{W,V}} \frac{b_i^2\Vert h_{\mathbf{v}_{0:i}}(\mathbf{X})\Vert_2^2}{\epsilon_i^2} \log(2W^2) \biggr] \\
    \le \, &\sum_{i=0}^K \bigl[ \sup_{{\mathbf{w}}_{-i}, \mathbf{v}\in\mathcal{W,V}} \frac{L_{\mathbf{w}}^2\log(2W^2)}{\lambda^2\epsilon^2} \frac{{b'_i}^2}{{\epsilon'_i}^2}  \bigr] \\
    &\; +   \sum_{i=0}^L \biggl[ \frac{4b_i^2(B+\frac{\prod_{i=0}^L \gamma_i a'_i}{\lambda})^2\prod_{i=0}^K \xi_i^2 a_i^2}{\epsilon^2}\sum_{i=0}^K\frac{b_i^2}{\alpha_i^2 a_i^2} \biggr] \\
    \le \, &\frac{\log(2W^2)\prod_{i=0}^L \gamma^2_i {a'_i}^2}{\lambda^2\epsilon^2} \sum_{k=0}^K \bigl[   \frac{{b'_i}^2(\sum_{i=k}^L\prod_{j=0}^i \gamma_i a_i)^2}{{\alpha'_i}^2{a'_i}^2}  \bigr] \\
    &\; +   \sum_{i=0}^L \biggl[ \frac{4b_i^2(B+\frac{\prod_{i=0}^L \gamma_i a'_i}{\lambda})^2\prod_{i=0}^K \xi_i^2 a_i^2}{\epsilon^2}\sum_{i=0}^K\frac{b_i^2}{\alpha_i^2 a_i^2} \biggr] \\
    \le \, &\frac{\log(2W^2)\prod_{i=0}^L \gamma^2_i {a'_i}^2 (\sum_{i=0}^L\prod_{j=0}^i \gamma_i a_i)^2}{\lambda^2\epsilon^2} \sum_{k=0}^K \bigl[   \frac{{b'_k}^2}{{\alpha'_k}^2{a'_k}^2}  \bigr] \\
    &\; +   \sum_{i=0}^L \biggl[ \frac{4b_i^2(B+\frac{\prod_{i=0}^L \gamma_i a'_i}{\lambda})^2\prod_{j=0}^K \xi_j^2 a_j^2}{\epsilon^2}\sum_{k=0}^K\frac{b_k^2}{\alpha_k^2 a_k^2} \biggr] 
    \\
    \le \, &\frac{4\log(2W^2)\prod_{i=0}^L \gamma^2_i {a'_i}^2 \bigl(\sum_{i=0}^L\prod_{j=0}^i \gamma_j a_j\bigr)^2}{\lambda^2 \epsilon^2} \sum_{i=0}^L \bigl[   \frac{{b'_i}^2}{{\alpha'_i}^2{a'_i}^2}  \bigr] \\
    &\; +  \frac{4\log(2W^2)(B+\frac{\prod_{i=0}^L \gamma_i a'_i}{\lambda})^2\prod_{i=0}^K \xi_i^2 a_i^2}{\epsilon^2} \sum_{i=0}^K\bigl[\frac{b_i^2}{\alpha_i^2 a_i^2} \bigr] \\
    = \, & \frac{C}{\epsilon^2}
\end{align*}
where we define
\begin{align*}
    C \, := \, 4\log(2W^2) \biggl[ \, &\frac{\prod_{i=0}^L \gamma^2_i {a'_i}^2 \bigl(\sum_{i=0}^L\prod_{j=0}^i \gamma_j a_j\bigr)^2}{\lambda^2} \left[\sum_{i=0}^L\frac{{b'_i}^{2/3}}{{a'_i}^{2/3}} \right]^{3} \\
    &+ \bigl(B+\frac{\prod_{i=0}^L \gamma_i a'_i}{\lambda}\bigr)^2\prod_{i=0}^K \xi_i^2 a_i^2\left[\sum_{i=0}^K\frac{b_i^{2/3}}{a_i^{2/3}} \right]^{3}\biggr].
\end{align*}

Now, based on the above covering-number bound, we use the Dudley entropy integral bound \cite{bartlett2017spectrally} which bounds the empirical Rademacher complexity of $\mathcal{F}\circ \Delta_{\mathcal{H}}|_S$ as 
\begin{align*}
    \mathcal{R}(\mathcal{F}\circ \Delta_{\mathcal{H}}|_S) \, &\le \, \inf_{\alpha\ge 0}\; \biggl\{\frac{4\alpha}{\sqrt{n}} + \frac{12}{n}\int_{\alpha}^{\sqrt{n}}\sqrt{\log \mathcal{N}\bigl(\mathcal{F}\circ\Delta_H|_S , \Vert\cdot\Vert_2,\epsilon \bigr)} d\epsilon \biggr\} \\
    & \le \, \inf_{\alpha\ge 0}\;\biggl\{\frac{4\alpha}{\sqrt{n}} + \log(\frac{\sqrt{n}}{\alpha} \frac{12\sqrt{C}}{n}) \\
    & \le \frac{4}{n^{3/2}} + \frac{18\log(n)\sqrt{C}}{n}
\end{align*}
where the last line follows from choosing $\alpha = 1/n$. Also, note that since for every non-negative constants $a,b\ge 0$ we have $\sqrt{a+b}\le \sqrt{a}+\sqrt{b}$, then
\begin{align*}
    \sqrt{C} \, \le \, 2\sqrt{\log(2D^2)} \biggl[ \, &\frac{\prod_{i=0}^L \gamma_i {a'_i} \bigl(\sum_{i=0}^L\prod_{j=0}^i \gamma_j a_j\bigr)}{\lambda} \left[\sum_{i=0}^L\frac{{b'_i}^{2/3}}{{a'_i}^{2/3}} \right]^{3/2} \\
    &+ \bigl(B+\frac{\prod_{i=0}^L \gamma_i a'_i}{\lambda}\bigr)\prod_{i=0}^K \xi_i a_i\left[\sum_{i=0}^K\frac{b_i^{2/3}}{a_i^{2/3}} \right]^{3/2}\biggr].
\end{align*}
Consequently, we have  the following bound where $R_{\mathcal{V}}$ and $R_{\mathcal{W}}$ are defined as in Theorem \ref{Thm: generalization bound}:
\begin{equation*}
     \mathcal{R}(\mathcal{F}\circ \Delta_{\mathcal{H}}|_S) \le \, \mathcal{O}\biggl(\frac{(B+\frac{L_{\mathbf{w}}}{\lambda}) \bigl(R_{\mathcal{V}}+\frac{1}{\tau^2}L_{\mathbf{w}}R_{\mathcal{W}}\bigr) \log(n) }{n}\log(D) \biggr) 
\end{equation*}
Therefore, according to standard Rademacher complexity-based generalization bounds \cite{bartlett2002rademacher}, for every $\omega>0$ with probability at least $1-\omega$  we have for every $\mathbf{v,w}\in\mathcal{V,W}$:
\begin{align*}
     &\frac{1}{n}\sum_{i=1}^n\left[ \ell\bigl(f_\mathbf{v}(\mathbf{x}_i+\delta^*_{\mathbf{w}}(\mathbf{x}_i,y_i)),y_i\bigr) \right] -  \mathbb{E}\left[ \ell\bigl(f_\mathbf{v}(\mathbf{X}+\delta^*_{\mathbf{w}}(\mathbf{X},Y)),Y\bigr)\right]  \\
     \le\, & \mathcal{O}\biggl(c\sqrt{\frac{\log(1/\omega)}{n}} + \frac{(B+\frac{L_{\mathbf{w}}}{\lambda}) \bigl(R_{\mathcal{V}}+\frac{1}{\tau^2}L_{\mathbf{w}}R_{\mathbf{w}}\bigr) \log(n) }{n}\log(D) \biggr),
\end{align*}
which implies that 
\begin{align*}
     \epsilon_{\mathrm{gen}}(\delta^*_{\mathbf{w}}) \, &= \,
     \min_{\mathbf{v}\in\mathcal{V}}\biggl\{\frac{1}{n}\sum_{i=1}^n\left[ \ell\bigl(f_\mathbf{v}(\mathbf{x}_i+\delta^*_{\mathbf{w}}(\mathbf{x}_i,y_i)),y_i\bigr) \right]\biggr\} -  \min_{\mathbf{v}\in\mathcal{V}}\biggl\{\mathbb{E}\left[ \ell\bigl(f_\mathbf{v}(\mathbf{X}+\delta^*_{\mathbf{w}}(\mathbf{X},Y)),Y\bigr)\right]\biggr\} \\
     &\le \, \max_{\mathbf{v}\in\mathcal{V}}\biggl\{\frac{1}{n}\sum_{i=1}^n\left[ \ell\bigl(f_\mathbf{v}(\mathbf{x}_i+\delta^*_{\mathbf{w}}(\mathbf{x}_i,y_i)),y_i\bigr) \right] -  \mathbb{E}\left[ \ell\bigl(f_\mathbf{v}(\mathbf{X}+\delta^*_{\mathbf{w}}(\mathbf{X},Y)),Y\bigr)\right] \biggr\} \\
     &\le \, \mathcal{O}\biggl(c\sqrt{\frac{\log(1/\omega)}{n}} + \frac{(B+\frac{L_{\mathbf{w}}}{\lambda}) \bigl(R_{\mathcal{V}}+\frac{1}{\tau^2}L_{\mathbf{w}}R_{\mathbf{w}}\bigr) \log(n) }{n}\log(D) \biggr).
\end{align*}
Therefore, the theorem's proof is complete.
\section{Additional Numerical Experiments}
% \usepackage{longtable}

% \usepackage{multirow}
%\subsection{Additional numerical results}
In this section, we report the complete set of our numerical results.
Table~\ref{tab:full_eraly_stopping} shows the complete results for DNNs regularized by early stopping, i.e. the complete version of Table~\ref{tab:early_stopping}. Moreover, Tables~\ref{tab:more_gen_trans1},\ref{tab:more_gen_trans2},\ref{tab:more_gen_trans3} are the full versions of Table~\ref{tab:generalization_transferability_SN}, demonstrating the relationship between generalization errors and transferability rates of the discussed datasets and DNN architectures. In addition to our earlier results, the accuracies for adversarially-perturbed training and test samples are also included in the tables.

%It might be concerned that the improvement of test adversarial example's transferability doesn't benefit from better generalization but from higher test accuracy. To eliminate this concern, we used the intersection of test samples which can be labeled correctly by the substitute models with and without regularization to evaluate transferability. 

In addition to the previous results, we also present the transferability rates averaged over test samples whose adversarial examples are predicted correctly by both the regularized and unregularized substitute DNNs. The transferability rates over those test samples intersecting the correctly predicted samples by regularized and unregularized substitute DNNs are shown in Tables~\ref{tab:full_eraly_stopping},\ref{tab:more_gen_trans1},\ref{tab:more_gen_trans2}, and \ref{tab:more_gen_trans3} under the title Transferability Rate-Int. Our numerical results suggest that over the test samples correctly predicted by both regularzied and unregularized DNNs, a better generalization score again results in higher transferability rates for designed adversarial examples. 

% Last, Figure \ref{fig:transferability_with_error_bar} presents the plots of the main text's Figure \ref{fig:tran_L2_FGM} with error bars.

\begin{table}[htbp]

\caption{Generalization and transferability rates for different DNN architectures and image datasets with and without early stopping (ES)}
\centering
\resizebox{\linewidth}{!}{\begin{tabular}{|cccccccc|}
\hline
Dataset & Model & Method & Train Acc & Test Acc & Gen.Err. & \begin{tabular}[c]{@{}c@{}}Transferability\\ Rate(VGG16)\end{tabular} & \begin{tabular}[c]{@{}c@{}}Transferability\\ Rate(ResNet18)\end{tabular} \\ \hline
\multicolumn{1}{|c|}{} & \multicolumn{1}{c|}{} & \multicolumn{1}{c|}{PGM} & 0.970 & 0.453 & 0.517 & 0.127 & 0.104 \\
\multicolumn{1}{|c|}{} & \multicolumn{1}{c|}{} & \multicolumn{1}{c|}{\cellcolor[HTML]{EFEFEF}PGM-ES} & \cellcolor[HTML]{EFEFEF}0.591 & \cellcolor[HTML]{EFEFEF}0.518 & \cellcolor[HTML]{EFEFEF}0.073 & \cellcolor[HTML]{EFEFEF}0.198 & \cellcolor[HTML]{EFEFEF}0.172 \\
\multicolumn{1}{|c|}{} & \multicolumn{1}{c|}{} & \multicolumn{1}{c|}{FGM} & 0.997 & 0.529 & 0.467 & 0.100 & 0.089 \\
\multicolumn{1}{|c|}{} & \multicolumn{1}{c|}{\multirow{-4}{*}{Inception}} & \multicolumn{1}{c|}{\cellcolor[HTML]{EFEFEF}FGM-ES} & \cellcolor[HTML]{EFEFEF}0.657 & \cellcolor[HTML]{EFEFEF}0.530 & \cellcolor[HTML]{EFEFEF}0.126 & \cellcolor[HTML]{EFEFEF}0.170 & \cellcolor[HTML]{EFEFEF}0.147 \\ \cline{2-8} 
\multicolumn{1}{|c|}{} & \multicolumn{1}{c|}{} & \multicolumn{1}{c|}{PGM} & 1.000 & 0.421 & 0.579 & 0.098 & 0.077 \\
\multicolumn{1}{|c|}{} & \multicolumn{1}{c|}{} & \multicolumn{1}{c|}{\cellcolor[HTML]{EFEFEF}PGM-ES} & \cellcolor[HTML]{EFEFEF}0.548 & \cellcolor[HTML]{EFEFEF}0.487 & \cellcolor[HTML]{EFEFEF}0.061 & \cellcolor[HTML]{EFEFEF}0.154 & \cellcolor[HTML]{EFEFEF}0.136 \\
\multicolumn{1}{|c|}{} & \multicolumn{1}{c|}{} & \multicolumn{1}{c|}{FGM} & 1.000 & 0.480 & 0.520 & 0.100 & 0.087 \\
\multicolumn{1}{|c|}{\multirow{-8}{*}{Cifar10}} & \multicolumn{1}{c|}{\multirow{-4}{*}{Alexnet}} & \multicolumn{1}{c|}{\cellcolor[HTML]{EFEFEF}FGM-ES} & \cellcolor[HTML]{EFEFEF}0.594 & \cellcolor[HTML]{EFEFEF}0.501 & \cellcolor[HTML]{EFEFEF}0.092 & \cellcolor[HTML]{EFEFEF}0.152 & \cellcolor[HTML]{EFEFEF}0.127 \\ \hline
\multicolumn{1}{|c|}{} & \multicolumn{1}{c|}{} & \multicolumn{1}{c|}{PGM} & 0.877 & 0.231 & 0.646 & 0.283 & 0.258 \\
\multicolumn{1}{|c|}{} & \multicolumn{1}{c|}{} & \multicolumn{1}{c|}{\cellcolor[HTML]{EFEFEF}PGM-ES} & \cellcolor[HTML]{EFEFEF}0.408 & \cellcolor[HTML]{EFEFEF}0.271 & \cellcolor[HTML]{EFEFEF}0.137 & \cellcolor[HTML]{EFEFEF}0.330 & \cellcolor[HTML]{EFEFEF}0.286 \\
\multicolumn{1}{|c|}{} & \multicolumn{1}{c|}{} & \multicolumn{1}{c|}{FGM} & 0.984 & 0.272 & 0.711 & 0.270 & 0.239 \\
\multicolumn{1}{|c|}{} & \multicolumn{1}{c|}{\multirow{-4}{*}{Inception}} & \multicolumn{1}{c|}{\cellcolor[HTML]{EFEFEF}FGM-ES} & \cellcolor[HTML]{EFEFEF}0.457 & \cellcolor[HTML]{EFEFEF}0.312 & \cellcolor[HTML]{EFEFEF}0.146 & \cellcolor[HTML]{EFEFEF}0.327 & \cellcolor[HTML]{EFEFEF}0.289 \\ \cline{2-8} 
\multicolumn{1}{|c|}{} & \multicolumn{1}{c|}{} & \multicolumn{1}{c|}{PGM} & 0.966 & 0.202 & 0.764 & 0.252 & 0.227 \\
\multicolumn{1}{|c|}{} & \multicolumn{1}{c|}{} & \multicolumn{1}{c|}{\cellcolor[HTML]{EFEFEF}PGM-ES} & \cellcolor[HTML]{EFEFEF}0.338 & \cellcolor[HTML]{EFEFEF}0.248 & \cellcolor[HTML]{EFEFEF}0.091 & \cellcolor[HTML]{EFEFEF}0.294 & \cellcolor[HTML]{EFEFEF}0.266 \\
\multicolumn{1}{|c|}{} & \multicolumn{1}{c|}{} & \multicolumn{1}{c|}{FGM} & 0.990 & 0.234 & 0.756 & 0.261 & 0.232 \\
\multicolumn{1}{|c|}{\multirow{-8}{*}{Cifar100}} & \multicolumn{1}{c|}{\multirow{-4}{*}{Alexnet}} & \multicolumn{1}{c|}{\cellcolor[HTML]{EFEFEF}FGM-ES} & \cellcolor[HTML]{EFEFEF}0.399 & \cellcolor[HTML]{EFEFEF}0.278 & \cellcolor[HTML]{EFEFEF}0.122 & \cellcolor[HTML]{EFEFEF}0.291 & \cellcolor[HTML]{EFEFEF}0.259 \\ \hline
\multicolumn{1}{|c|}{} & \multicolumn{1}{c|}{} & \multicolumn{1}{c|}{PGM} & 0.925 & 0.585 & 0.341 & 0.207 & 0.220 \\
\multicolumn{1}{|c|}{} & \multicolumn{1}{c|}{} & \multicolumn{1}{c|}{\cellcolor[HTML]{EFEFEF}PGM-ES} & \cellcolor[HTML]{EFEFEF}0.711 & \cellcolor[HTML]{EFEFEF}0.654 & \cellcolor[HTML]{EFEFEF}0.057 & \cellcolor[HTML]{EFEFEF}0.298 & \cellcolor[HTML]{EFEFEF}0.322 \\
\multicolumn{1}{|c|}{} & \multicolumn{1}{c|}{} & \multicolumn{1}{c|}{FGM} & 0.998 & 0.619 & 0.380 & 0.136 & 0.129 \\
\multicolumn{1}{|c|}{} & \multicolumn{1}{c|}{\multirow{-4}{*}{Inception}} & \multicolumn{1}{c|}{\cellcolor[HTML]{EFEFEF}FGM-ES} & \cellcolor[HTML]{EFEFEF}0.898 & \cellcolor[HTML]{EFEFEF}0.718 & \cellcolor[HTML]{EFEFEF}0.180 & \cellcolor[HTML]{EFEFEF}0.213 & \cellcolor[HTML]{EFEFEF}0.219 \\ \cline{2-8} 
\multicolumn{1}{|c|}{} & \multicolumn{1}{c|}{} & \multicolumn{1}{c|}{PGM} & 0.848 & 0.541 & 0.307 & 0.211 & 0.228 \\
\multicolumn{1}{|c|}{} & \multicolumn{1}{c|}{} & \multicolumn{1}{c|}{\cellcolor[HTML]{EFEFEF}PGM-ES} & \cellcolor[HTML]{EFEFEF}0.624 & \cellcolor[HTML]{EFEFEF}0.594 & \cellcolor[HTML]{EFEFEF}0.030 & \cellcolor[HTML]{EFEFEF}0.256 & \cellcolor[HTML]{EFEFEF}0.278 \\
\multicolumn{1}{|c|}{} & \multicolumn{1}{c|}{} & \multicolumn{1}{c|}{FGM} & 0.949 & 0.576 & 0.373 & 0.157 & 0.170 \\
\multicolumn{1}{|c|}{\multirow{-8}{*}{SVHN}} & \multicolumn{1}{c|}{\multirow{-4}{*}{Alexnet}} & \multicolumn{1}{c|}{\cellcolor[HTML]{EFEFEF}FGM-ES} & \cellcolor[HTML]{EFEFEF}0.691 & \cellcolor[HTML]{EFEFEF}0.627 & \cellcolor[HTML]{EFEFEF}0.064 & \cellcolor[HTML]{EFEFEF}0.241 & \cellcolor[HTML]{EFEFEF}0.260 \\ \hline
\end{tabular}}
\label{tab:full_eraly_stopping}
\end{table}

\begin{table}[htbp]
\caption{Generalization and transferability rates for different DNN architectures and image datasets with and without early stopping (ES), Transferability Rate-Int. means averaged transferability rate on adversarial examples correctly labeled by both the regularized and unregularized DNNs.}
\centering
\resizebox{0.8\linewidth}{!}{
\begin{tabular}{|cccccc|}
\hline
Dataset & Model & Method & Gen.Err. & \begin{tabular}[c]{@{}c@{}}Transferability\\ Rate-Int.(VGG16)\end{tabular} & \begin{tabular}[c]{@{}c@{}}Transferability\\ Rate-Int.(ResNet18)\end{tabular} \\ \hline
\multicolumn{1}{|c|}{} & \multicolumn{1}{c|}{} & \multicolumn{1}{c|}{PGM} & 0.517 & 0.030 & 0.127 \\
\multicolumn{1}{|c|}{} & \multicolumn{1}{c|}{} & \multicolumn{1}{c|}{\cellcolor[HTML]{EFEFEF}PGM-ES} & \cellcolor[HTML]{EFEFEF}0.073 & \cellcolor[HTML]{EFEFEF}0.063 & \cellcolor[HTML]{EFEFEF}0.198 \\
\multicolumn{1}{|c|}{} & \multicolumn{1}{c|}{} & \multicolumn{1}{c|}{FGM} & 0.467 & 0.032 & 0.100 \\
\multicolumn{1}{|c|}{} & \multicolumn{1}{c|}{\multirow{-4}{*}{Inception}} & \multicolumn{1}{c|}{\cellcolor[HTML]{EFEFEF}FGM-ES} & \cellcolor[HTML]{EFEFEF}0.126 & \cellcolor[HTML]{EFEFEF}0.077 & \cellcolor[HTML]{EFEFEF}0.170 \\ \cline{2-6} 
\multicolumn{1}{|c|}{} & \multicolumn{1}{c|}{} & \multicolumn{1}{c|}{PGM} & 0.579 & 0.037 & 0.098 \\
\multicolumn{1}{|c|}{} & \multicolumn{1}{c|}{} & \multicolumn{1}{c|}{\cellcolor[HTML]{EFEFEF}PGM-ES} & \cellcolor[HTML]{EFEFEF}0.061 & \cellcolor[HTML]{EFEFEF}0.077 & \cellcolor[HTML]{EFEFEF}0.154 \\
\multicolumn{1}{|c|}{} & \multicolumn{1}{c|}{} & \multicolumn{1}{c|}{FGM} & 0.520 & 0.035 & 0.100 \\
\multicolumn{1}{|c|}{\multirow{-8}{*}{Cifar10}} & \multicolumn{1}{c|}{\multirow{-4}{*}{Alexnet}} & \multicolumn{1}{c|}{\cellcolor[HTML]{EFEFEF}FGM-ES} & \cellcolor[HTML]{EFEFEF}0.092 & \cellcolor[HTML]{EFEFEF}0.076 & \cellcolor[HTML]{EFEFEF}0.152 \\ \hline
\multicolumn{1}{|c|}{} & \multicolumn{1}{c|}{} & \multicolumn{1}{c|}{PGM} & 0.646 & 0.141 & 0.283 \\
\multicolumn{1}{|c|}{} & \multicolumn{1}{c|}{} & \multicolumn{1}{c|}{\cellcolor[HTML]{EFEFEF}PGM-ES} & \cellcolor[HTML]{EFEFEF}0.137 & \cellcolor[HTML]{EFEFEF}0.160 & \cellcolor[HTML]{EFEFEF}0.330 \\
\multicolumn{1}{|c|}{} & \multicolumn{1}{c|}{} & \multicolumn{1}{c|}{FGM} & 0.711 & 0.126 & 0.270 \\
\multicolumn{1}{|c|}{} & \multicolumn{1}{c|}{\multirow{-4}{*}{Inception}} &\multicolumn{1}{c|}{FGM-ES}  & \cellcolor[HTML]{EFEFEF}0.146 & \cellcolor[HTML]{EFEFEF}0.154 & \cellcolor[HTML]{EFEFEF}0.327 \\ \cline{2-6} 
\multicolumn{1}{|c|}{} & \multicolumn{1}{c|}{} & \multicolumn{1}{c|}{PGM} & 0.764 & 0.114 & 0.252 \\
\multicolumn{1}{|c|}{} & \multicolumn{1}{c|}{} & \multicolumn{1}{c|}{\cellcolor[HTML]{EFEFEF}PGM-ES} & \cellcolor[HTML]{EFEFEF}0.091 & \cellcolor[HTML]{EFEFEF}0.159 & \cellcolor[HTML]{EFEFEF}0.294 \\
\multicolumn{1}{|c|}{} & \multicolumn{1}{c|}{} & \multicolumn{1}{c|}{FGM} & 0.756 & 0.127 & 0.261 \\
\multicolumn{1}{|c|}{\multirow{-8}{*}{Cifar100}} & \multicolumn{1}{c|}{\multirow{-4}{*}{Alexnet}} & \multicolumn{1}{c|}{\cellcolor[HTML]{EFEFEF}FGM-ES} & \cellcolor[HTML]{EFEFEF}0.122 & \cellcolor[HTML]{EFEFEF}0.180 & \cellcolor[HTML]{EFEFEF}0.291 \\ \hline
\multicolumn{1}{|c|}{} & \multicolumn{1}{c|}{} & \multicolumn{1}{c|}{PGM} & 0.341 & 0.024 & 0.207 \\
\multicolumn{1}{|c|}{} & \multicolumn{1}{c|}{} & \multicolumn{1}{c|}{\cellcolor[HTML]{EFEFEF}PGM-ES} & \cellcolor[HTML]{EFEFEF}0.057 & \cellcolor[HTML]{EFEFEF}0.102 & \cellcolor[HTML]{EFEFEF}0.298 \\
\multicolumn{1}{|c|}{} & \multicolumn{1}{c|}{} & \multicolumn{1}{c|}{FGM} & 0.380 & 0.010 & 0.136 \\
\multicolumn{1}{|c|}{} & \multicolumn{1}{c|}{\multirow{-4}{*}{Inception}} & \multicolumn{1}{c|}{\cellcolor[HTML]{EFEFEF}FGM-ES} & \cellcolor[HTML]{EFEFEF}0.180 & \cellcolor[HTML]{EFEFEF}0.103 & \cellcolor[HTML]{EFEFEF}0.213 \\ \cline{2-6} 
\multicolumn{1}{|c|}{} & \multicolumn{1}{c|}{} & \multicolumn{1}{c|}{PGM} & 0.307 & 0.044 & 0.211 \\
\multicolumn{1}{|c|}{} & \multicolumn{1}{c|}{} & \multicolumn{1}{c|}{\cellcolor[HTML]{EFEFEF}PGM-ES} & \cellcolor[HTML]{EFEFEF}0.030 & \cellcolor[HTML]{EFEFEF}0.080 & \cellcolor[HTML]{EFEFEF}0.256 \\
\multicolumn{1}{|c|}{} & \multicolumn{1}{c|}{} & \multicolumn{1}{c|}{FGM} & 0.373 & 0.013 & 0.157 \\
\multicolumn{1}{|c|}{\multirow{-8}{*}{SVHN}} & \multicolumn{1}{c|}{\multirow{-4}{*}{Alexnet}} & \multicolumn{1}{c|}{\cellcolor[HTML]{EFEFEF}FGM-ES} & \cellcolor[HTML]{EFEFEF}0.064 & \cellcolor[HTML]{EFEFEF}0.092 & \cellcolor[HTML]{EFEFEF}0.241 \\ \hline
\end{tabular}}
\label{tab:early_stopping_intersection}
\end{table}

\begin{table}[htbp]
\centering
% \vspace{-10mm}
\caption{Generalization and transferability rates on CIFAR-10 data, with and without using spectral regularization}
\resizebox{\linewidth}{!}{

\begin{tabular}{|ccccccccc|}
\hline
Dataset & Model & Method & $\beta$ & Train Acc & Test Acc & Gen.Err. & \begin{tabular}[c]{@{}c@{}}Transferability\\ Rate(VGG16)\end{tabular} & \begin{tabular}[c]{@{}c@{}}Transferability\\ Rate(ResNet18)\end{tabular} \\ \hline
\multicolumn{1}{|c|}{} & \multicolumn{1}{c|}{} & \multicolumn{1}{c|}{} & \multicolumn{1}{c|}{$\infty$} & 0.998 & 0.532 & 0.466 & 0.092 & 0.078 \\
\multicolumn{1}{|c|}{} & \multicolumn{1}{c|}{} & \multicolumn{1}{c|}{} & \multicolumn{1}{c|}{\cellcolor[HTML]{EFEFEF}1} & \cellcolor[HTML]{EFEFEF}0.830 & \cellcolor[HTML]{EFEFEF}0.511 & \cellcolor[HTML]{EFEFEF}0.320 & \cellcolor[HTML]{EFEFEF}0.136 & \cellcolor[HTML]{EFEFEF}0.113 \\
\multicolumn{1}{|c|}{} & \multicolumn{1}{c|}{} & \multicolumn{1}{c|}{} & \multicolumn{1}{c|}{1.3} & 0.936 & 0.479 & 0.456 & 0.124 & 0.097 \\
\multicolumn{1}{|c|}{} & \multicolumn{1}{c|}{} & \multicolumn{1}{c|}{} & \multicolumn{1}{c|}{1.6} & 0.997 & 0.509 & 0.488 & 0.099 & 0.084 \\
\multicolumn{1}{|c|}{} & \multicolumn{1}{c|}{} & \multicolumn{1}{c|}{} & \multicolumn{1}{c|}{2} & 0.999 & 0.541 & 0.458 & 0.087 & 0.073 \\
\multicolumn{1}{|c|}{} & \multicolumn{1}{c|}{} & \multicolumn{1}{c|}{\multirow{-6}{*}{FGM($L_2$)}} & \multicolumn{1}{c|}{3} & 1.000 & 0.554 & 0.445 & 0.090 & 0.077 \\ \cline{3-4}
\multicolumn{1}{|c|}{} & \multicolumn{1}{c|}{} & \multicolumn{1}{c|}{} & \multicolumn{1}{c|}{$\infty$} & 0.951 & 0.443 & 0.508 & 0.104 & 0.084 \\
\multicolumn{1}{|c|}{} & \multicolumn{1}{c|}{} & \multicolumn{1}{c|}{} & \multicolumn{1}{c|}{\cellcolor[HTML]{EFEFEF}1} & \cellcolor[HTML]{EFEFEF}0.759 & \cellcolor[HTML]{EFEFEF}0.502 & \cellcolor[HTML]{EFEFEF}0.258 & \cellcolor[HTML]{EFEFEF}0.150 & \cellcolor[HTML]{EFEFEF}0.134 \\
\multicolumn{1}{|c|}{} & \multicolumn{1}{c|}{} & \multicolumn{1}{c|}{} & \multicolumn{1}{c|}{1.3} & 0.913 & 0.459 & 0.454 & 0.125 & 0.099 \\
\multicolumn{1}{|c|}{} & \multicolumn{1}{c|}{} & \multicolumn{1}{c|}{} & \multicolumn{1}{c|}{1.6} & 0.945 & 0.447 & 0.499 & 0.115 & 0.095 \\
\multicolumn{1}{|c|}{} & \multicolumn{1}{c|}{} & \multicolumn{1}{c|}{} & \multicolumn{1}{c|}{2} & 0.979 & 0.451 & 0.529 & 0.110 & 0.092 \\
\multicolumn{1}{|c|}{} & \multicolumn{1}{c|}{} & \multicolumn{1}{c|}{\multirow{-6}{*}{PGM($L_2$)}} & \multicolumn{1}{c|}{3} & 0.988 & 0.468 & 0.520 & 0.103 & 0.082 \\ \cline{3-4}
\multicolumn{1}{|c|}{} & \multicolumn{1}{c|}{} & \multicolumn{1}{c|}{} & \multicolumn{1}{c|}{$\infty$} & 0.946 & 0.442 & 0.504 & 0.124 & 0.101 \\
\multicolumn{1}{|c|}{} & \multicolumn{1}{c|}{} & \multicolumn{1}{c|}{} & \multicolumn{1}{c|}{\cellcolor[HTML]{EFEFEF}1} & \cellcolor[HTML]{EFEFEF}0.694 & \cellcolor[HTML]{EFEFEF}0.451 & \cellcolor[HTML]{EFEFEF}0.243 & \cellcolor[HTML]{EFEFEF}0.191 & \cellcolor[HTML]{EFEFEF}0.153 \\
\multicolumn{1}{|c|}{} & \multicolumn{1}{c|}{} & \multicolumn{1}{c|}{} & \multicolumn{1}{c|}{1.3} & 0.842 & 0.421 & 0.421 & 0.165 & 0.129 \\
\multicolumn{1}{|c|}{} & \multicolumn{1}{c|}{} & \multicolumn{1}{c|}{} & \multicolumn{1}{c|}{1.6} & 0.951 & 0.419 & 0.531 & 0.129 & 0.106 \\
\multicolumn{1}{|c|}{} & \multicolumn{1}{c|}{} & \multicolumn{1}{c|}{} & \multicolumn{1}{c|}{2} & 0.980 & 0.453 & 0.527 & 0.121 & 0.103 \\
\multicolumn{1}{|c|}{} & \multicolumn{1}{c|}{} & \multicolumn{1}{c|}{\multirow{-6}{*}{FGM($L_\infty$)}} & \multicolumn{1}{c|}{3} & 0.988 & 0.470 & 0.517 & 0.112 & 0.091 \\ \cline{3-4}
\multicolumn{1}{|c|}{} & \multicolumn{1}{c|}{} & \multicolumn{1}{c|}{} & \multicolumn{1}{c|}{$\infty$} & 0.708 & 0.525 & 0.184 & 0.419 & 0.393 \\
\multicolumn{1}{|c|}{} & \multicolumn{1}{c|}{} & \multicolumn{1}{c|}{} & \multicolumn{1}{c|}{\cellcolor[HTML]{EFEFEF}1} & \cellcolor[HTML]{EFEFEF}0.627 & \cellcolor[HTML]{EFEFEF}0.545 & \cellcolor[HTML]{EFEFEF}0.082 & \cellcolor[HTML]{EFEFEF}0.497 & \cellcolor[HTML]{EFEFEF}0.474 \\
\multicolumn{1}{|c|}{} & \multicolumn{1}{c|}{} & \multicolumn{1}{c|}{} & \multicolumn{1}{c|}{1.3} & 0.674 & 0.534 & 0.140 & 0.466 & 0.443 \\
\multicolumn{1}{|c|}{} & \multicolumn{1}{c|}{} & \multicolumn{1}{c|}{} & \multicolumn{1}{c|}{1.6} & 0.710 & 0.520 & 0.190 & 0.442 & 0.422 \\
\multicolumn{1}{|c|}{} & \multicolumn{1}{c|}{} & \multicolumn{1}{c|}{} & \multicolumn{1}{c|}{2} & 0.739 & 0.505 & 0.234 & 0.408 & 0.389 \\
\multicolumn{1}{|c|}{} & \multicolumn{1}{c|}{\multirow{-24}{*}{Inception}} & \multicolumn{1}{c|}{\multirow{-6}{*}{PGM($L_\infty$)}} & \multicolumn{1}{c|}{3} & 0.745 & 0.504 & 0.241 & 0.398 & 0.380 \\ \cline{2-9} 
\multicolumn{1}{|c|}{} & \multicolumn{1}{c|}{} & \multicolumn{1}{c|}{} & \multicolumn{1}{c|}{$\infty$} & 1.000 & 0.495 & 0.505 & 0.089 & 0.070 \\
\multicolumn{1}{|c|}{} & \multicolumn{1}{c|}{} & \multicolumn{1}{c|}{} & \multicolumn{1}{c|}{\cellcolor[HTML]{EFEFEF}1} & \cellcolor[HTML]{EFEFEF}0.977 & \cellcolor[HTML]{EFEFEF}0.526 & \cellcolor[HTML]{EFEFEF}0.451 & \cellcolor[HTML]{EFEFEF}0.147 & \cellcolor[HTML]{EFEFEF}0.122 \\
\multicolumn{1}{|c|}{} & \multicolumn{1}{c|}{} & \multicolumn{1}{c|}{} & \multicolumn{1}{c|}{1.3} & 1.000 & 0.496 & 0.504 & 0.132 & 0.105 \\
\multicolumn{1}{|c|}{} & \multicolumn{1}{c|}{} & \multicolumn{1}{c|}{} & \multicolumn{1}{c|}{1.6} & 1.000 & 0.474 & 0.526 & 0.115 & 0.090 \\
\multicolumn{1}{|c|}{} & \multicolumn{1}{c|}{} & \multicolumn{1}{c|}{} & \multicolumn{1}{c|}{2} & 1.000 & 0.477 & 0.523 & 0.108 & 0.088 \\
\multicolumn{1}{|c|}{} & \multicolumn{1}{c|}{} & \multicolumn{1}{c|}{\multirow{-6}{*}{FGM($L_2$)}} & \multicolumn{1}{c|}{3} & 1.000 & 0.496 & 0.504 & 0.098 & 0.079 \\ \cline{3-4}
\multicolumn{1}{|c|}{} & \multicolumn{1}{c|}{} & \multicolumn{1}{c|}{} & \multicolumn{1}{c|}{$\infty$} & 0.999 & 0.454 & 0.545 & 0.105 & 0.087 \\
\multicolumn{1}{|c|}{} & \multicolumn{1}{c|}{} & \multicolumn{1}{c|}{\multirow{-2}{*}{PGM($L_2$)}} & \multicolumn{1}{c|}{\cellcolor[HTML]{EFEFEF}1} & \cellcolor[HTML]{EFEFEF}0.861 & \cellcolor[HTML]{EFEFEF}0.519 & \cellcolor[HTML]{EFEFEF}0.342 & \cellcolor[HTML]{EFEFEF}0.162 & \cellcolor[HTML]{EFEFEF}0.139 \\ \cline{3-4}
\multicolumn{1}{|c|}{} & \multicolumn{1}{c|}{} & \multicolumn{1}{c|}{} & \multicolumn{1}{c|}{$\infty$} & 0.996 & 0.400 & 0.596 & 0.102 & 0.089 \\
\multicolumn{1}{|c|}{} & \multicolumn{1}{c|}{} & \multicolumn{1}{c|}{} & \multicolumn{1}{c|}{\cellcolor[HTML]{EFEFEF}1} & \cellcolor[HTML]{EFEFEF}0.914 & \cellcolor[HTML]{EFEFEF}0.426 & \cellcolor[HTML]{EFEFEF}0.487 & \cellcolor[HTML]{EFEFEF}0.199 & \cellcolor[HTML]{EFEFEF}0.169 \\
\multicolumn{1}{|c|}{} & \multicolumn{1}{c|}{} & \multicolumn{1}{c|}{} & \multicolumn{1}{c|}{1.3} & 0.998 & 0.387 & 0.610 & 0.169 & 0.142 \\
\multicolumn{1}{|c|}{} & \multicolumn{1}{c|}{} & \multicolumn{1}{c|}{} & \multicolumn{1}{c|}{1.6} & 1.000 & 0.368 & 0.632 & 0.146 & 0.120 \\
\multicolumn{1}{|c|}{} & \multicolumn{1}{c|}{} & \multicolumn{1}{c|}{} & \multicolumn{1}{c|}{2} & 1.000 & 0.420 & 0.580 & 0.128 & 0.106 \\
\multicolumn{1}{|c|}{} & \multicolumn{1}{c|}{} & \multicolumn{1}{c|}{\multirow{-6}{*}{FGM($L_\infty$)}} & \multicolumn{1}{c|}{3} & 1.000 & 0.435 & 0.565 & 0.115 & 0.096 \\ \cline{3-4}
\multicolumn{1}{|c|}{} & \multicolumn{1}{c|}{} & \multicolumn{1}{c|}{} & \multicolumn{1}{c|}{$\infty$} & 0.685 & 0.474 & 0.211 & 0.446 & 0.423 \\
\multicolumn{1}{|c|}{} & \multicolumn{1}{c|}{} & \multicolumn{1}{c|}{} & \multicolumn{1}{c|}{\cellcolor[HTML]{EFEFEF}1} & \cellcolor[HTML]{EFEFEF}0.628 & \cellcolor[HTML]{EFEFEF}0.473 & \cellcolor[HTML]{EFEFEF}0.156 & \cellcolor[HTML]{EFEFEF}0.472 & \cellcolor[HTML]{EFEFEF}0.457 \\
\multicolumn{1}{|c|}{} & \multicolumn{1}{c|}{} & \multicolumn{1}{c|}{} & \multicolumn{1}{c|}{1.3} & 0.680 & 0.436 & 0.245 & 0.441 & 0.423 \\
\multicolumn{1}{|c|}{} & \multicolumn{1}{c|}{} & \multicolumn{1}{c|}{} & \multicolumn{1}{c|}{1.6} & 0.697 & 0.420 & 0.277 & 0.421 & 0.403 \\
\multicolumn{1}{|c|}{} & \multicolumn{1}{c|}{} & \multicolumn{1}{c|}{} & \multicolumn{1}{c|}{2} & 0.679 & 0.408 & 0.271 & 0.404 & 0.388 \\
\multicolumn{1}{|c|}{\multirow{-44}{*}{Cifar10}} & \multicolumn{1}{c|}{\multirow{-20}{*}{Alexnet}} & \multicolumn{1}{c|}{\multirow{-6}{*}{PGM($L_\infty$)}} & \multicolumn{1}{c|}{3} & 0.628 & 0.439 & 0.189 & 0.426 & 0.405 \\ \hline
\end{tabular}

}
\label{tab:more_gen_trans1}
\end{table}

\begin{table}[htbp]
\centering
% \vspace{-10mm}
\caption{Generalization and transferability rates on CIFAR-100 dataset, with and without spectral regularization.}
\resizebox{\linewidth}{!}{
\begin{tabular}{|ccccccccc|}
\hline
Dataset & Model & Method & $\beta$ & Train Acc & Test Acc & Gen.Err. & \begin{tabular}[c]{@{}c@{}}Transferability\\ Rate(VGG16)\end{tabular} & \begin{tabular}[c]{@{}c@{}}Transferability\\ Rate(ResNet18)\end{tabular} \\ \hline
\multicolumn{1}{|c|}{} & \multicolumn{1}{c|}{} & \multicolumn{1}{c|}{} & \multicolumn{1}{c|}{$\infty$} & 0.996 & 0.279 & 0.717 & 0.268 & 0.236 \\
\multicolumn{1}{|c|}{} & \multicolumn{1}{c|}{} & \multicolumn{1}{c|}{} & \multicolumn{1}{c|}{1} & 0.761 & 0.277 & 0.484 & 0.250 & 0.290 \\
\multicolumn{1}{|c|}{} & \multicolumn{1}{c|}{} & \multicolumn{1}{c|}{} & \multicolumn{1}{c|}{\cellcolor[HTML]{EFEFEF}1.3} & \cellcolor[HTML]{EFEFEF}{\color[HTML]{000000} 0.853} & \cellcolor[HTML]{EFEFEF}{\color[HTML]{000000} 0.294} & \cellcolor[HTML]{EFEFEF}{\color[HTML]{000000} 0.558} & \cellcolor[HTML]{EFEFEF}{\color[HTML]{000000} 0.313} & \cellcolor[HTML]{EFEFEF}{\color[HTML]{000000} 0.275} \\
\multicolumn{1}{|c|}{} & \multicolumn{1}{c|}{} & \multicolumn{1}{c|}{} & \multicolumn{1}{c|}{1.6} & 0.998 & 0.260 & 0.738 & 0.236 & 0.263 \\
\multicolumn{1}{|c|}{} & \multicolumn{1}{c|}{} & \multicolumn{1}{c|}{} & \multicolumn{1}{c|}{2} & 1.000 & 0.284 & 0.715 & 0.238 & 0.272 \\
\multicolumn{1}{|c|}{} & \multicolumn{1}{c|}{} & \multicolumn{1}{c|}{\multirow{-6}{*}{FGM($L_2$)}} & \multicolumn{1}{c|}{3} & 0.999 & 0.262 & 0.736 & 0.221 & 0.257 \\ \cline{3-4}
\multicolumn{1}{|c|}{} & \multicolumn{1}{c|}{} & \multicolumn{1}{c|}{} & \multicolumn{1}{c|}{$\infty$} & 0.857 & 0.255 & 0.602 & 0.303 & 0.270 \\
\multicolumn{1}{|c|}{} & \multicolumn{1}{c|}{} & \multicolumn{1}{c|}{} & \multicolumn{1}{c|}{1} & 0.604 & 0.255 & 0.349 & 0.329 & 0.297 \\
\multicolumn{1}{|c|}{} & \multicolumn{1}{c|}{} & \multicolumn{1}{c|}{} & \multicolumn{1}{c|}{\cellcolor[HTML]{EFEFEF}1.3} & \cellcolor[HTML]{EFEFEF}0.750 & \cellcolor[HTML]{EFEFEF}0.256 & \cellcolor[HTML]{EFEFEF}0.494 & \cellcolor[HTML]{EFEFEF}0.330 & \cellcolor[HTML]{EFEFEF}0.301 \\
\multicolumn{1}{|c|}{} & \multicolumn{1}{c|}{} & \multicolumn{1}{c|}{} & \multicolumn{1}{c|}{1.6} & 0.893 & 0.230 & 0.664 & 0.305 & 0.265 \\
\multicolumn{1}{|c|}{} & \multicolumn{1}{c|}{} & \multicolumn{1}{c|}{} & \multicolumn{1}{c|}{2} & 0.868 & 0.233 & 0.635 & 0.307 & 0.277 \\
\multicolumn{1}{|c|}{} & \multicolumn{1}{c|}{\multirow{-12}{*}{Inception}} & \multicolumn{1}{c|}{\multirow{-6}{*}{PGM($L_2$)}} & \multicolumn{1}{c|}{3} & 0.948 & 0.220 & 0.728 & 0.290 & 0.248 \\ \cline{2-9} 
\multicolumn{1}{|c|}{} & \multicolumn{1}{c|}{} & \multicolumn{1}{c|}{} & \multicolumn{1}{c|}{$\infty$} & 1.000 & 0.242 & 0.758 & 0.258 & 0.232 \\
\multicolumn{1}{|c|}{} & \multicolumn{1}{c|}{} & \multicolumn{1}{c|}{} & \multicolumn{1}{c|}{\cellcolor[HTML]{EFEFEF}1} & \cellcolor[HTML]{EFEFEF}0.925 & \cellcolor[HTML]{EFEFEF}0.315 & \cellcolor[HTML]{EFEFEF}0.611 & \cellcolor[HTML]{EFEFEF}0.342 & \cellcolor[HTML]{EFEFEF}0.310 \\
\multicolumn{1}{|c|}{} & \multicolumn{1}{c|}{} & \multicolumn{1}{c|}{} & \multicolumn{1}{c|}{1.3} & 1.000 & 0.289 & 0.710 & 0.304 & 0.266 \\
\multicolumn{1}{|c|}{} & \multicolumn{1}{c|}{} & \multicolumn{1}{c|}{} & \multicolumn{1}{c|}{1.6} & 1.000 & 0.285 & 0.715 & 0.291 & 0.248 \\
\multicolumn{1}{|c|}{} & \multicolumn{1}{c|}{} & \multicolumn{1}{c|}{} & \multicolumn{1}{c|}{2} & 1.000 & 0.284 & 0.716 & 0.288 & 0.253 \\
\multicolumn{1}{|c|}{} & \multicolumn{1}{c|}{} & \multicolumn{1}{c|}{\multirow{-6}{*}{FGM($L_2$)}} & \multicolumn{1}{c|}{3} & 1.000 & 0.269 & 0.730 & 0.265 & 0.240 \\ \cline{3-4}
\multicolumn{1}{|c|}{} & \multicolumn{1}{c|}{} & \multicolumn{1}{c|}{} & \multicolumn{1}{c|}{$\infty$} & 1.000 & 0.210 & 0.789 & 0.229 & 0.260 \\
\multicolumn{1}{|c|}{} & \multicolumn{1}{c|}{} & \multicolumn{1}{c|}{} & \multicolumn{1}{c|}{\cellcolor[HTML]{EFEFEF}1} & \cellcolor[HTML]{EFEFEF}0.889 & \cellcolor[HTML]{EFEFEF}0.288 & \cellcolor[HTML]{EFEFEF}0.601 & \cellcolor[HTML]{EFEFEF}0.323 & \cellcolor[HTML]{EFEFEF}0.353 \\
\multicolumn{1}{|c|}{} & \multicolumn{1}{c|}{} & \multicolumn{1}{c|}{} & \multicolumn{1}{c|}{1.3} & 0.998 & 0.234 & 0.763 & 0.307 & 0.282 \\
\multicolumn{1}{|c|}{} & \multicolumn{1}{c|}{} & \multicolumn{1}{c|}{} & \multicolumn{1}{c|}{1.6} & 1.000 & 0.227 & 0.773 & 0.306 & 0.261 \\
\multicolumn{1}{|c|}{} & \multicolumn{1}{c|}{} & \multicolumn{1}{c|}{} & \multicolumn{1}{c|}{2} & 1.000 & 0.229 & 0.771 & 0.283 & 0.259 \\
\multicolumn{1}{|c|}{\multirow{-24}{*}{Cifar100}} & \multicolumn{1}{c|}{\multirow{-12}{*}{Alexnet}} & \multicolumn{1}{c|}{\multirow{-6}{*}{PGM($L_2$)}} & \multicolumn{1}{c|}{3} & 1.000 & 0.213 & 0.787 & 0.272 & 0.238 \\ \hline
\end{tabular}
}
\label{tab:more_gen_trans2}
\end{table}

\begin{table}[htbp]
\centering
% \vspace{-10mm}
\caption{Generalization and transferability rates on SVHN dataset, with and without spectral regularization.}
\resizebox{\linewidth}{!}{
\begin{tabular}{|ccccccccc|}
\hline
Dataset & Model & Method & $\beta$ & Train Acc & Test Acc & Gen.Err. & \begin{tabular}[c]{@{}c@{}}Transferability\\ Rate(VGG16)\end{tabular} & \begin{tabular}[c]{@{}c@{}}Transferability\\ Rate(ResNet18)\end{tabular} \\ \hline
\multicolumn{1}{|c|}{} & \multicolumn{1}{c|}{} & \multicolumn{1}{c|}{} & \multicolumn{1}{c|}{$\infty$} & 0.991 & 0.618 & 0.373 & 0.134 & 0.126 \\
\multicolumn{1}{|c|}{} & \multicolumn{1}{c|}{} & \multicolumn{1}{c|}{} & \multicolumn{1}{c|}{\cellcolor[HTML]{EFEFEF}1} & \cellcolor[HTML]{EFEFEF}0.848 & \cellcolor[HTML]{EFEFEF}0.645 & \cellcolor[HTML]{EFEFEF}0.203 & \cellcolor[HTML]{EFEFEF}0.277 & \cellcolor[HTML]{EFEFEF}0.257 \\
\multicolumn{1}{|c|}{} & \multicolumn{1}{c|}{} & \multicolumn{1}{c|}{} & \multicolumn{1}{c|}{1.3} & 0.967 & 0.605 & 0.362 & 0.223 & 0.209 \\
\multicolumn{1}{|c|}{} & \multicolumn{1}{c|}{} & \multicolumn{1}{c|}{} & \multicolumn{1}{c|}{1.6} & 0.998 & 0.573 & 0.426 & 0.202 & 0.187 \\
\multicolumn{1}{|c|}{} & \multicolumn{1}{c|}{} & \multicolumn{1}{c|}{} & \multicolumn{1}{c|}{2} & 1.000 & 0.571 & 0.429 & 0.158 & 0.149 \\
\multicolumn{1}{|c|}{} & \multicolumn{1}{c|}{} & \multicolumn{1}{c|}{\multirow{-6}{*}{FGM($L_2$)}} & \multicolumn{1}{c|}{3} & 1.000 & 0.642 & 0.358 & 0.121 & 0.118 \\ \cline{3-4}
\multicolumn{1}{|c|}{} & \multicolumn{1}{c|}{} & \multicolumn{1}{c|}{} & \multicolumn{1}{c|}{$\infty$} & 0.708 & 0.524 & 0.184 & 0.393 & 0.419 \\
\multicolumn{1}{|c|}{} & \multicolumn{1}{c|}{} & \multicolumn{1}{c|}{} & \multicolumn{1}{c|}{\cellcolor[HTML]{EFEFEF}1} & \cellcolor[HTML]{EFEFEF}0.627 & \cellcolor[HTML]{EFEFEF}0.545 & \cellcolor[HTML]{EFEFEF}0.082 & \cellcolor[HTML]{EFEFEF}0.474 & \cellcolor[HTML]{EFEFEF}0.497 \\
\multicolumn{1}{|c|}{} & \multicolumn{1}{c|}{} & \multicolumn{1}{c|}{} & \multicolumn{1}{c|}{1.3} & 0.674 & 0.534 & 0.140 & 0.466 & 0.443 \\
\multicolumn{1}{|c|}{} & \multicolumn{1}{c|}{} & \multicolumn{1}{c|}{} & \multicolumn{1}{c|}{1.6} & 0.710 & 0.520 & 0.190 & 0.442 & 0.422 \\
\multicolumn{1}{|c|}{} & \multicolumn{1}{c|}{} & \multicolumn{1}{c|}{} & \multicolumn{1}{c|}{2} & 0.739 & 0.505 & 0.234 & 0.408 & 0.389 \\
\multicolumn{1}{|c|}{} & \multicolumn{1}{c|}{\multirow{-12}{*}{Inception}} & \multicolumn{1}{c|}{\multirow{-6}{*}{PGM($L_2$)}} & \multicolumn{1}{c|}{3} & 0.745 & 0.504 & 0.241 & 0.398 & 0.380 \\ \cline{2-9} 
\multicolumn{1}{|c|}{} & \multicolumn{1}{c|}{} & \multicolumn{1}{c|}{} & \multicolumn{1}{c|}{$\infty$} & 0.991 & 0.618 & 0.373 & 0.134 & 0.126 \\
\multicolumn{1}{|c|}{} & \multicolumn{1}{c|}{} & \multicolumn{1}{c|}{} & \multicolumn{1}{c|}{\cellcolor[HTML]{EFEFEF}1} & \cellcolor[HTML]{EFEFEF}0.848 & \cellcolor[HTML]{EFEFEF}0.645 & \cellcolor[HTML]{EFEFEF}0.203 & \cellcolor[HTML]{EFEFEF}0.277 & \cellcolor[HTML]{EFEFEF}0.257 \\
\multicolumn{1}{|c|}{} & \multicolumn{1}{c|}{} & \multicolumn{1}{c|}{} & \multicolumn{1}{c|}{1.3} & 0.967 & 0.605 & 0.362 & 0.223 & 0.209 \\
\multicolumn{1}{|c|}{} & \multicolumn{1}{c|}{} & \multicolumn{1}{c|}{} & \multicolumn{1}{c|}{1.6} & 0.998 & 0.573 & 0.426 & 0.202 & 0.187 \\
\multicolumn{1}{|c|}{} & \multicolumn{1}{c|}{} & \multicolumn{1}{c|}{} & \multicolumn{1}{c|}{2} & 1.000 & 0.571 & 0.429 & 0.158 & 0.149 \\
\multicolumn{1}{|c|}{} & \multicolumn{1}{c|}{} & \multicolumn{1}{c|}{\multirow{-6}{*}{FGM($L_2$)}} & \multicolumn{1}{c|}{3} & 0.999 & 0.581 & 0.418 & 0.145 & 0.133 \\ \cline{3-4}
\multicolumn{1}{|c|}{} & \multicolumn{1}{c|}{} & \multicolumn{1}{c|}{} & \multicolumn{1}{c|}{$\infty$} & 0.844 & 0.546 & 0.298 & 0.211 & 0.225 \\
\multicolumn{1}{|c|}{} & \multicolumn{1}{c|}{} & \multicolumn{1}{c|}{} & \multicolumn{1}{c|}{\cellcolor[HTML]{EFEFEF}1} & \cellcolor[HTML]{EFEFEF}0.817 & \cellcolor[HTML]{EFEFEF}0.618 & \cellcolor[HTML]{EFEFEF}0.199 & \cellcolor[HTML]{EFEFEF}0.276 & \cellcolor[HTML]{EFEFEF}0.292 \\
\multicolumn{1}{|c|}{} & \multicolumn{1}{c|}{} & \multicolumn{1}{c|}{} & \multicolumn{1}{c|}{1.3} & 0.936 & 0.570 & 0.366 & 0.229 & 0.244 \\
\multicolumn{1}{|c|}{} & \multicolumn{1}{c|}{} & \multicolumn{1}{c|}{} & \multicolumn{1}{c|}{1.6} & 0.976 & 0.539 & 0.437 & 0.197 & 0.209 \\
\multicolumn{1}{|c|}{} & \multicolumn{1}{c|}{} & \multicolumn{1}{c|}{} & \multicolumn{1}{c|}{2} & 0.992 & 0.520 & 0.473 & 0.172 & 0.182 \\
\multicolumn{1}{|c|}{\multirow{-24}{*}{SVHN}} & \multicolumn{1}{c|}{\multirow{-12}{*}{Alexnet}} & \multicolumn{1}{c|}{\multirow{-6}{*}{PGM($L_2$)}} & \multicolumn{1}{c|}{3} & 0.986 & 0.511 & 0.476 & 0.150 & 0.155 \\ \hline
\end{tabular}}
\label{tab:more_gen_trans3}
\end{table}

\begin{table}[htbp]
\centering
% \vspace{-10mm}
\caption{Generalization and transferability rates for different DNN architectures and image datasets with and without spectral regularization. Transferability Rate-Int. means averaged transferability rate on adversarial examples correctly labeled by both the regularized and unregularized DNNs.}
\resizebox{0.8\linewidth}{!}{
\begin{tabular}{|ccccccc|}
\hline
Dataset & Model & Method & $\beta$ & Gen.Err. & \begin{tabular}[c]{@{}c@{}}Transferability\\ Rate-Int(VGG16)\end{tabular} & \begin{tabular}[c]{@{}c@{}}Transferability\\ Rate-Int(ResNet18)\end{tabular} \\ \hline
\multicolumn{1}{|c|}{} & \multicolumn{1}{c|}{} & \multicolumn{1}{c|}{} & \multicolumn{1}{c|}{$\infty$} & 0.466 & 0.032 & 0.026 \\
\multicolumn{1}{|c|}{} & \multicolumn{1}{c|}{} & \multicolumn{1}{c|}{\multirow{-2}{*}{FGM($L_2$)}} & \multicolumn{1}{c|}{\cellcolor[HTML]{EFEFEF}1} & \cellcolor[HTML]{EFEFEF}0.320 & \cellcolor[HTML]{EFEFEF}0.077 & \cellcolor[HTML]{EFEFEF}0.058 \\ \cline{3-4}
\multicolumn{1}{|c|}{} & \multicolumn{1}{c|}{} & \multicolumn{1}{c|}{} & \multicolumn{1}{c|}{$\infty$} & 0.508 & 0.030 & 0.026 \\
\multicolumn{1}{|c|}{} & \multicolumn{1}{c|}{} & \multicolumn{1}{c|}{\multirow{-2}{*}{PGM($L_2$)}} & \multicolumn{1}{c|}{\cellcolor[HTML]{EFEFEF}1} & \cellcolor[HTML]{EFEFEF}0.258 & \cellcolor[HTML]{EFEFEF}0.063 & \cellcolor[HTML]{EFEFEF}0.052 \\ \cline{3-4}
\multicolumn{1}{|c|}{} & \multicolumn{1}{c|}{} & \multicolumn{1}{c|}{} & \multicolumn{1}{c|}{$\infty$} & 0.504 & 0.029 & 0.028 \\
\multicolumn{1}{|c|}{} & \multicolumn{1}{c|}{} & \multicolumn{1}{c|}{\multirow{-2}{*}{FGM($L_\infty$)}} & \multicolumn{1}{c|}{\cellcolor[HTML]{EFEFEF}1} & \cellcolor[HTML]{EFEFEF}0.243 & \cellcolor[HTML]{EFEFEF}0.070 & \cellcolor[HTML]{EFEFEF}0.090 \\ \cline{3-4}
\multicolumn{1}{|c|}{} & \multicolumn{1}{c|}{} & \multicolumn{1}{c|}{} & \multicolumn{1}{c|}{$\infty$} & 0.184 & 0.136 & 0.181 \\
\multicolumn{1}{|c|}{} & \multicolumn{1}{c|}{\multirow{-8}{*}{Inception}} & \multicolumn{1}{c|}{\multirow{-2}{*}{PGM($L_\infty$)}} & \multicolumn{1}{c|}{\cellcolor[HTML]{EFEFEF}1} & \cellcolor[HTML]{EFEFEF}0.082 & \cellcolor[HTML]{EFEFEF}0.182 & \cellcolor[HTML]{EFEFEF}0.162 \\ \cline{2-7} 
\multicolumn{1}{|c|}{} & \multicolumn{1}{c|}{} & \multicolumn{1}{c|}{} & \multicolumn{1}{c|}{$\infty$} & 0.505 & 0.035 & 0.029 \\
\multicolumn{1}{|c|}{} & \multicolumn{1}{c|}{} & \multicolumn{1}{c|}{\multirow{-2}{*}{FGM($L_2$)}} & \multicolumn{1}{c|}{\cellcolor[HTML]{EFEFEF}1} & \cellcolor[HTML]{EFEFEF}0.451 & \cellcolor[HTML]{EFEFEF}0.076 & \cellcolor[HTML]{EFEFEF}0.068 \\ \cline{3-4}
\multicolumn{1}{|c|}{} & \multicolumn{1}{c|}{} & \multicolumn{1}{c|}{} & \multicolumn{1}{c|}{$\infty$} & 0.545 & 0.037 & 0.030 \\
\multicolumn{1}{|c|}{} & \multicolumn{1}{c|}{} & \multicolumn{1}{c|}{\multirow{-2}{*}{PGM($L_2$)}} & \multicolumn{1}{c|}{\cellcolor[HTML]{EFEFEF}1} & \cellcolor[HTML]{EFEFEF}0.342 & \cellcolor[HTML]{EFEFEF}0.077 & \cellcolor[HTML]{EFEFEF}0.063 \\ \cline{3-4}
\multicolumn{1}{|c|}{} & \multicolumn{1}{c|}{} & \multicolumn{1}{c|}{} & \multicolumn{1}{c|}{$\infty$} & 0.596 & 0.039 & 0.019 \\
\multicolumn{1}{|c|}{} & \multicolumn{1}{c|}{} & \multicolumn{1}{c|}{\multirow{-2}{*}{FGM($L_\infty$)}} & \multicolumn{1}{c|}{\cellcolor[HTML]{EFEFEF}1} & \cellcolor[HTML]{EFEFEF}0.487 & \cellcolor[HTML]{EFEFEF}0.089 & \cellcolor[HTML]{EFEFEF}0.070 \\ \cline{3-4}
\multicolumn{1}{|c|}{} & \multicolumn{1}{c|}{} & \multicolumn{1}{c|}{} & \multicolumn{1}{c|}{$\infty$} & 0.211 & 0.227 & 0.222 \\
\multicolumn{1}{|c|}{\multirow{-16}{*}{Cifar10}} & \multicolumn{1}{c|}{\multirow{-8}{*}{Alexnet}} & \multicolumn{1}{c|}{\multirow{-2}{*}{PGM($L_\infty$)}} & \multicolumn{1}{c|}{\cellcolor[HTML]{EFEFEF}1} & \cellcolor[HTML]{EFEFEF}0.156 & \cellcolor[HTML]{EFEFEF}0.271 & \cellcolor[HTML]{EFEFEF}0.248 \\ \hline
\multicolumn{1}{|c|}{} & \multicolumn{1}{c|}{} & \multicolumn{1}{c|}{} & \multicolumn{1}{c|}{$\infty$} & 0.717 & 0.126 & 0.112 \\
\multicolumn{1}{|c|}{} & \multicolumn{1}{c|}{} & \multicolumn{1}{c|}{\multirow{-2}{*}{FGM($L_2$)}} & \multicolumn{1}{c|}{\cellcolor[HTML]{EFEFEF}1.3} & \cellcolor[HTML]{EFEFEF}0.558 & \cellcolor[HTML]{EFEFEF}0.154 & \cellcolor[HTML]{EFEFEF}0.131 \\ \cline{3-4}
\multicolumn{1}{|c|}{} & \multicolumn{1}{c|}{} & \multicolumn{1}{c|}{} & \multicolumn{1}{c|}{$\infty$} & 0.602 & 0.141 & 0.123 \\
\multicolumn{1}{|c|}{} & \multicolumn{1}{c|}{\multirow{-4}{*}{Inception}} & \multicolumn{1}{c|}{\multirow{-2}{*}{PGM($L_2$)}} & \multicolumn{1}{c|}{\cellcolor[HTML]{EFEFEF}1.3} & \cellcolor[HTML]{EFEFEF}0.494 & \cellcolor[HTML]{EFEFEF}0.160 & \cellcolor[HTML]{EFEFEF}0.141 \\ \cline{2-7} 
\multicolumn{1}{|c|}{} & \multicolumn{1}{c|}{} & \multicolumn{1}{c|}{} & \multicolumn{1}{c|}{$\infty$} & 0.758 & 0.127 & 0.101 \\
\multicolumn{1}{|c|}{} & \multicolumn{1}{c|}{} & \multicolumn{1}{c|}{\multirow{-2}{*}{FGM($L_2$)}} & \multicolumn{1}{c|}{\cellcolor[HTML]{EFEFEF}1} & \cellcolor[HTML]{EFEFEF}0.611 & \cellcolor[HTML]{EFEFEF}0.180 & \cellcolor[HTML]{EFEFEF}0.159 \\ \cline{3-4}
\multicolumn{1}{|c|}{} & \multicolumn{1}{c|}{} & \multicolumn{1}{c|}{} & \multicolumn{1}{c|}{$\infty$} & 0.789 & 0.114 & 0.108 \\
\multicolumn{1}{|c|}{\multirow{-8}{*}{Cifar100}} & \multicolumn{1}{c|}{\multirow{-4}{*}{Alexnet}} & \multicolumn{1}{c|}{\multirow{-2}{*}{PGM($L_2$)}} & \multicolumn{1}{c|}{\cellcolor[HTML]{EFEFEF}1} & \cellcolor[HTML]{EFEFEF}0.601 & \cellcolor[HTML]{EFEFEF}0.159 & \cellcolor[HTML]{EFEFEF}0.151 \\ \hline
\multicolumn{1}{|c|}{} & \multicolumn{1}{c|}{} & \multicolumn{1}{c|}{} & \multicolumn{1}{c|}{$\infty$} & 0.373 & 0.010 & 0.013 \\
\multicolumn{1}{|c|}{} & \multicolumn{1}{c|}{} & \multicolumn{1}{c|}{\multirow{-2}{*}{FGM($L_2$)}} & \multicolumn{1}{c|}{\cellcolor[HTML]{EFEFEF}1} & \cellcolor[HTML]{EFEFEF}0.203 & \cellcolor[HTML]{EFEFEF}0.103 & \cellcolor[HTML]{EFEFEF}0.125 \\ \cline{3-4}
\multicolumn{1}{|c|}{} & \multicolumn{1}{c|}{} & \multicolumn{1}{c|}{} & \multicolumn{1}{c|}{$\infty$} & 0.342 & 0.024 & 0.031 \\
\multicolumn{1}{|c|}{} & \multicolumn{1}{c|}{\multirow{-4}{*}{Inception}} & \multicolumn{1}{c|}{\multirow{-2}{*}{PGM($L_2$)}} & \multicolumn{1}{c|}{\cellcolor[HTML]{EFEFEF}1} & \cellcolor[HTML]{EFEFEF}0.115 & \cellcolor[HTML]{EFEFEF}0.102 & \cellcolor[HTML]{EFEFEF}0.125 \\ \cline{2-7} 
\multicolumn{1}{|c|}{} & \multicolumn{1}{c|}{} & \multicolumn{1}{c|}{} & \multicolumn{1}{c|}{$\infty$} & 0.373 & 0.013 & 0.014 \\
\multicolumn{1}{|c|}{} & \multicolumn{1}{c|}{} & \multicolumn{1}{c|}{\multirow{-2}{*}{FGM($L_2$)}} & \multicolumn{1}{c|}{\cellcolor[HTML]{EFEFEF}1} & \cellcolor[HTML]{EFEFEF}0.203 & \cellcolor[HTML]{EFEFEF}0.092 & \cellcolor[HTML]{EFEFEF}0.112 \\ \cline{3-4}
\multicolumn{1}{|c|}{} & \multicolumn{1}{c|}{} & \multicolumn{1}{c|}{} & \multicolumn{1}{c|}{$\infty$} & 0.298 & 0.044 & 0.053 \\
\multicolumn{1}{|c|}{\multirow{-8}{*}{SVHN}} & \multicolumn{1}{c|}{\multirow{-4}{*}{Alexnet}} & \multicolumn{1}{c|}{\multirow{-2}{*}{PGM($L_2$)}} & \multicolumn{1}{c|}{\cellcolor[HTML]{EFEFEF}1} & \cellcolor[HTML]{EFEFEF}0.199 & \cellcolor[HTML]{EFEFEF}0.080 & \cellcolor[HTML]{EFEFEF}0.101 \\ \hline
\end{tabular}
}
\label{tab:spectral_normalization_intersection}
\end{table}

% \begin{figure}[t]

%     \centering
%     \includegraphics[width=\linewidth]{Figures/Experiment_Fig/trans_3_datasets_only_test_witherrorbar.png}
%     \caption{Figure 1 including error bars}
%     \label{fig:transferability_with_error_bar}
% \end{figure}

\iffalse
\subsection{Numerical Results for ERM-trained substitute DNNs}
\begin{figure}[t]
    \centering
    \includegraphics[width=\linewidth]{NeurIPS2021/submission/Figures/Experiment_Fig/trans_ERM.png}
    \caption{Transferability for perturbations on ERM-trained models}
    \label{fig:ERM-trained_models}
    
    \centering
    \includegraphics[width=\linewidth]{NeurIPS2021/submission/Figures/Experiment_Fig/trans_3_datasets_only_test_witherrorbar.png}
    \caption{Figure with error bar}
    \label{fig:transferability_with_error_bar}
\end{figure}
\textcolor{red}{
Fig.~\ref{fig:ERM-trained_models} shows a draft of the attack performance of $L_2$-FGM on standard trained models. Attack error is the accuracy of the perturbations, and transferability rate follows the same definition as above. I'm not sure if this result helps because in some cases, the transferability rate gap between adversarial samples generated from training and test samples might still decrease with the increase of $\beta$. I also evaluated this by $L_2$-PGD, $L_\infty$-FGM and $L_\infty$-PGD, but this problems still exists.
}

\textcolor{red}{
Fig.~\ref{fig:transferability_with_error_bar} shows the same results as in Fig.~\ref{fig:tran_L2_FGM} with error bars.}
\fi

\end{appendices}

\end{document}